\PassOptionsToPackage{prologue,dvipsnames}{xcolor}
\documentclass[10pt,twocolumn,letterpaper]{article}
\usepackage{tcolorbox}
\usepackage[pagenumbers]{iccv} 
\usepackage{makecell}
\usepackage{bbding}
\usepackage{multirow}

\usepackage{amsmath,amsfonts,bm}

\def\eqref#1{equation~\ref{#1}}

\def\1{\bm{1}}

\DeclareMathAlphabet{\mathsfit}{\encodingdefault}{\sfdefault}{m}{sl}
\SetMathAlphabet{\mathsfit}{bold}{\encodingdefault}{\sfdefault}{bx}{n}

\definecolor{iccvblue}{rgb}{0.21,0.49,0.74}
\usepackage[pagebackref,breaklinks,colorlinks,allcolors=iccvblue]{hyperref}

\usepackage[ruled]{algorithm2e} %

\SetAlFnt{\small}
\SetAlCapFnt{\small}
\SetAlCapNameFnt{\small}
\SetAlCapHSkip{0pt}
\usepackage{pifont}
\usepackage{colortbl}
\usepackage{xspace}

\newlength\savewidth

\usepackage{listings}
\newcommand\blfootnote[1]{%
  \begingroup
  \renewcommand\thefootnote{}\footnote{#1}%
  \addtocounter{footnote}{-1}%
  \endgroup
}

\colorlet{punct}{red!60!black}
\definecolor{background}{HTML}{EEEEEE}
\definecolor{delim}{RGB}{20,105,176}
\colorlet{numb}{magenta!60!black}

\lstdefinelanguage{json}{
    basicstyle=\normalfont\ttfamily,
    numbers=left,
    numberstyle=\scriptsize,
    stepnumber=1,
    numbersep=8pt,
    showstringspaces=false,
    breaklines=true,
    frame=lines,
    backgroundcolor=\color{background},
    literate=
     *{0}{{{\color{numb}0}}}{1}
      {1}{{{\color{numb}1}}}{1}
      {2}{{{\color{numb}2}}}{1}
      {3}{{{\color{numb}3}}}{1}
      {4}{{{\color{numb}4}}}{1}
      {5}{{{\color{numb}5}}}{1}
      {6}{{{\color{numb}6}}}{1}
      {7}{{{\color{numb}7}}}{1}
      {8}{{{\color{numb}8}}}{1}
      {9}{{{\color{numb}9}}}{1}
      {:}{{{\color{punct}{:}}}}{1}
      {,}{{{\color{punct}{,}}}}{1}
      {\{}{{{\color{delim}{\{}}}}{1}
      {\}}{{{\color{delim}{\}}}}}{1}
      {[}{{{\color{delim}{[}}}}{1}
      {]}{{{\color{delim}{]}}}}{1},
}

\title{Multimodal Latent Diffusion Model for Complex Sewing Pattern Generation}

\author{\parbox{\textwidth}{\centering Shengqi Liu$^{1}$,
        Yuhao Cheng$^{1}$\thanks{Project leader}, Zhuo Chen$^{1}$, Xingyu Ren$^{1}$,  Wenhan Zhu$^{2}$, \\ Lincheng Li$^{3}$, Mengxiao Bi$^{3}$, Xiaokang Yang$^{1}$, Yichao Yan$^{1}$\thanks{Corresponding authors}
      }
         \\ \\
{\parbox{\textwidth}{\centering $^1$MoE Key Lab of Artificial Intelligence, AI Institute, Shanghai Jiao Tong University, China}} \\
{\parbox{\textwidth}{\centering $^2$Xueshen AI, $^3$NetEase Fuxi AI Lab}}
 }

\begin{document}
\twocolumn[{
\renewcommand\twocolumn[1][]{#1}
    \maketitle
    \begin{center}
        \includegraphics[width=\linewidth]{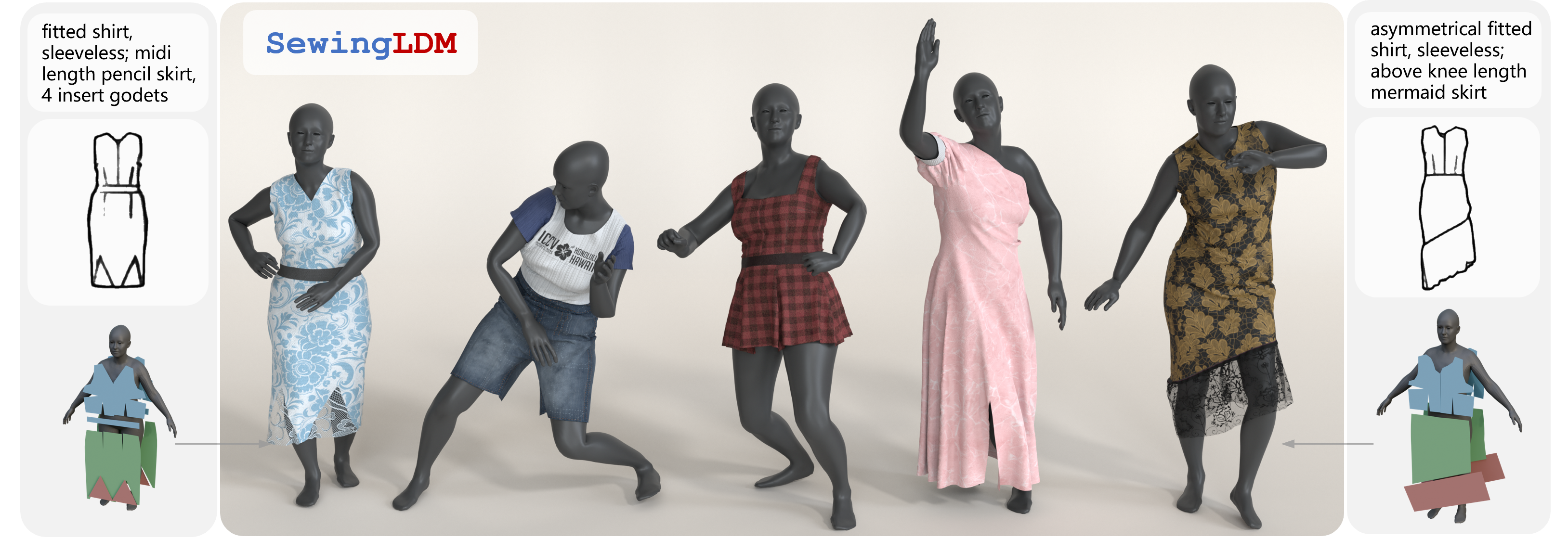}
        \captionof{figure}{ SewingLDM can generate complex sewing pattern designs under the condition of texts, garment sketches, and body shapes, demonstrating detailed control ability. The generated garments can be seamlessly integrated into the CG pipeline for simulation and animation, achieving vivid and photo-realistic rendering results.}
        \label{fig:teaser}
    \end{center}
}]

\begin{abstract}
Generating sewing patterns in garment design is receiving increasing attention due to its CG-friendly and flexible-editing nature.
Previous sewing pattern generation methods have been able to produce exquisite clothing, but struggle to design complex garments with detailed control.
To address these issues, we propose \textbf{SewingLDM}, a multi-modal generative model that generates sewing patterns controlled by text prompts, body shapes, and garment sketches.
Initially, we extend the original vector of sewing patterns into a more comprehensive representation to cover more intricate details and then compress them into a compact latent space.
To learn the sewing pattern distribution in the latent space, we design a two-step training strategy to inject the multi-modal conditions, \ie, body shapes, text prompts, and garment sketches, into a diffusion model, ensuring the generated garments are body-suited and detail-controlled.
Comprehensive qualitative and quantitative experiments show the effectiveness of our proposed method, significantly surpassing previous approaches in terms of complex garment design and various body adaptability.
Our project page: \href{https://shengqiliu1.github.io/SewingLDM}{https://shengqiliu1.github.io/SewingLDM}.
\end{abstract}    
\blfootnote{\textsuperscript{$*$} Project leader \textsuperscript{$\dag$}Corresponding author}
\section{Introduction}
\label{sec:intro}

Clothes play a pivotal role in shaping human aesthetics and physique, where appropriate clothes can beautify their overall appearance and highlight human physical attributes.
Therefore, garment design has always been a crucial component that
significantly impacts both digital character creation and real-life human visual presentation.
In recent years, many garment generation methods~\cite{sarafianos2024garment3dgen, li2024garmentdreamer,korosteleva2022neuraltailor, he2024dresscode, liu2023towards,korosteleva2023garmentcode,srivastava2024wordrobe,baldrati2023multimodal,morelli2022dresscode, morelli2023ladi,luo2024garverselod,chen2024anidress,xiang2023drivable,zheng2024physavatar,kim2024stableviton} have emerged to generate desired garments for users under various conditions.

Typically, garment generation can be classified into 2D and 3D methods.
The 2D generation methods~\cite{baldrati2023multimodal,morelli2022dresscode, morelli2023ladi,kim2024stableviton} can produce visually appealing results but cannot maintain consistency across different views, failing to drape on human bodies.
Therefore, many recent works are focusing on 3D cloth generation~\cite{sarafianos2024garment3dgen, li2024garmentdreamer}.
Although these methods can generate high-quality meshes or neural fields, they pose the challenge of clipping between clothes and bodies when draping clothes onto the human body, and they are incompatible with the digital garment production pipeline.
Meanwhile, the sewing pattern is a more widely used representation for garments in the industry because it facilitates both physical simulation and animation in CG-friendly fashions~\cite{maya, blender}.
Previous methods for sewing pattern generation~\cite{korosteleva2022neuraltailor, he2024dresscode, liu2023towards} achieve fantastic garment generation but fall short in designing complex features.
Most recently, concurrent works~\cite{bian2024chatgarment, nakayama2024aipparel, zhou2024design2garmentcode} introduce complex features with reduced tokenization, leading to significant advancements in garment generation.
However, these methods typically ignore human body shapes, preventing the creation of made-to-measure garments.
Apart from learning-based methods, the parametric garment design tool~\cite{korosteleva2023garmentcode} allows users to model complex garments and considers relations with body shapes.
Nonetheless, this tool requires pre-defined templates and detailed body measurements, and then with a delicate selection of control parameters to generate the garments, which require professional knowledge of garment designs.
In summary, there are two main challenges in suitable sewing pattern generation: 1) Designing a general representation for complex sewing patterns, and 2) Enabling control over garment details and ensuring garments are body-suited.

To address these issues, we design a novel architecture named \textbf{SewingLDM} for generating complex sewing patterns under the control of texts, body shapes, and garment sketches. 
To represent complex designs of sewing patterns, we especially design an extended representation to encompass intricate types of edges and attachments of garments, enabling more general and complex garment learning.
Subsequently, we train an auto-encoder model to compress the representation into a compact latent space, which not only enables efficient training of complex sewing pattern generation with fixed memory consumption but also ensures scalability for future applications involving more intricate garment designs.
To achieve multi-modal controlled and body-aware sewing pattern generation, we design a two-step training strategy to introduce the control signals into the latent diffusion model.
In the first step, we train the diffusion model under the condition of texts, serving as a coarse fundamental model for additional control signal injection.
In the second step, we further embed the knowledge of sketches and body shapes into the diffusion model by fusing the features after the first block and fine-tuning the output layers within the attention module, providing additional control of garment details and ensuring the generated garments fit various body shapes end to end without the need of delicate body measurements.
Based on the proposed framework, SewingLDM can generate complex garments that fit various body shapes and align with user-provided text descriptions or garment sketches.

Our generated sewing patterns can be seamlessly integrated into subsequent CG pipelines, facilitating editing and animation processes.
After simulation, the garment mesh can be combined with current texture generation methods~\cite{zhang2024dreammat,youwang2024paint,zhang2024fabricdiffusion,lopes2024material} or handcrafted texture to generate colored garments, as shown in~\cref{fig:teaser}, demonstrating our fantastic generation ability.
Comprehensive qualitative and quantitative experiments show the superiority of our proposed method in terms of complex garment design and various body adaptability compared with previous methods.
To summarize, our main contributions include:
\begin{itemize}
    \item We design a novel architecture, dubbed SewingLDM, for sewing pattern generation conditioned by texts, body shapes, and garment sketches, enabling precisely controlled and body-suited garment generation.
    \item We design an extended representation to cover complex sewing patterns and compress it into a compact latent space enabling the training of complex sewing patterns generation and maintaining high reconstruction quality.
    \item We design a two-step training strategy to better inject the multi-modal control signals into a diffusion model, yielding superior generation performance and controllability.
\end{itemize}

\section{Related Work}
\label{sec:related}

\begin{figure*}[t]
  \centering
  \includegraphics[width=0.95\linewidth]{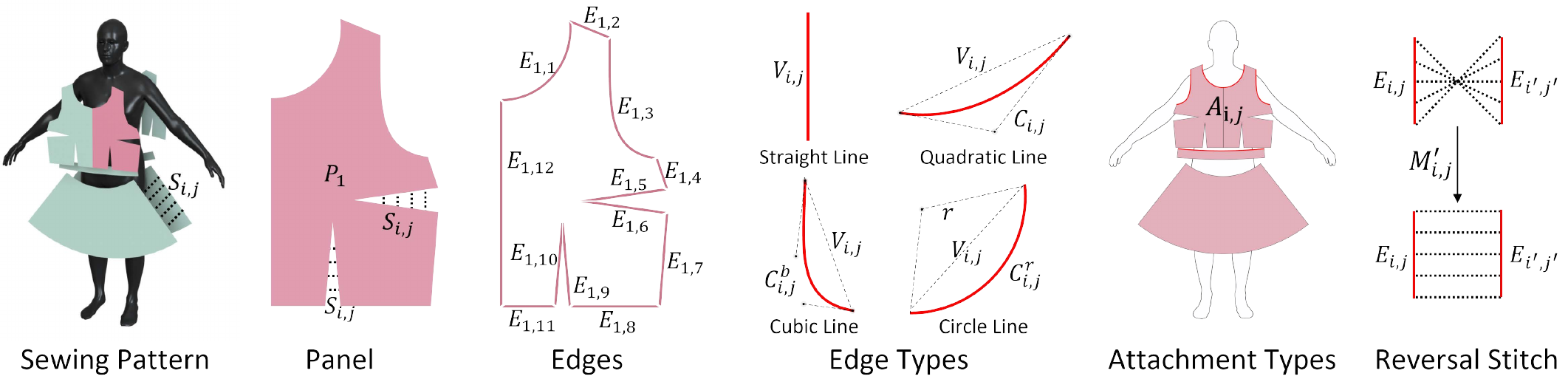}
    \vspace{-0.5em}
   \caption{\textbf{Sewing pattern.} Sewing patterns are CAD representations of garments, containing 2D shapes and 3D placement of cloth. They consist of panels, and panels consist of edges joined from beginning to end. Between panels or inner panels, stitches are used to connect edges to form clothes. For each edge, different kinds of lines are utilized to conform to body contours. Additionally, complex sewing patterns need additional attachment constraints during simulation for certain edges, which is highlighted in red in the attachment types region. Besides, for the stitch between panels and inner panels, a reversal stitch flag is sometimes needed to reverse the stitch direction.}
   \vspace{-1.5em}
   \label{fig:sewing_pattern_pre}
\end{figure*}

\noindent\textbf{Multimodal-guided 3D Generation.}
The advancements in large models~\cite{radford2021learning, rombach2022high, achiam2023gpt} have encouraged the emergence of recent 3D generation models~\cite{wang2022clip, jetchev2021clipmatrix, jain2022zero, michel2022text2mesh, chen2023fantasia3d, liu2024one, zhang2024clay, sun2024recent}, entering a new era of 3D generation.
Some models~\cite{DBLP:conf/iclr/PooleJBM23, tang2023make, zeng2023avatarbooth, cao2024dreamavatar,metzer2023latent,luo2024garverselod} focus on generating implicit neural radiance fields corresponding to the text description, whereas others~\cite{tang2023dreamgaussian, chen2023fantasia3d, liu2024directional} extend their ability to generate 3D meshes with BRDF materials.
Especially, some models~\cite{sarafianos2024garment3dgen, srivastava2024wordrobe, li2024garmentdreamer,luo2024garverselod} dive into human clothes generation.
Garment3DGen~\cite{sarafianos2024garment3dgen} can generate textured 3D mesh from images or text conditions by adjusting template meshes.
GarmentDreamer~\cite{li2024garmentdreamer} utilizes 3D Gaussian Splatting (GS)~\cite{kerbl20233d} as guidance to create 3D garment meshes from textual prompts.
WordRobe~\cite{srivastava2024wordrobe} leverages unsigned distance field (UDF)~\cite{guillard2022udf} to represent 3D garments and generate 3D garment meshes under text guidance.
GarVerseLOD~\cite{luo2024garverselod} proposes a hierarchical framework to recover different levels of garment details through single images.
Although these methods can generate visual-appealing garment meshes, their compatibility with CG pipelines remains a challenge, hindering seamless integration into modern industry workflows.
In contrast, our method aims at sewing pattern generation, facilitating utilization in CG processes.

\noindent\textbf{3D Sewing Pattern Modeling.}
Existing fashion CAD software tools, such as Clo3D~\cite{clo3d} and Marvelous Designer~\cite{marvelousdesigner}, allow users to edit sewing patterns and simulate desired cloth outcomes.
While these methods integrate the most advanced garment design, they heavily rely on artists to manually draw and adjust the shapes of sewing patterns, requiring a substantial of professional manual processing.
Consequently, ongoing studies are focusing on automating the adjustment of sewing patterns~\cite{Bartle2016c,Wang2009InteractiveCurves,Meng2012,Liu2018c,Wolff2023DesigningMovement,Pietroni2022,feng2024neural}, reconstructing sewing patterns~\cite{Jeong2015c,Yang2018c,Hasler2007ReverseGarments,Chen2015,Bang2021EstimatingData,korosteleva2022neuraltailor,liu2023towards}, and assisting with complex garment design~\cite{Li2018FoldSketch:Folds,Wang2018,Fondevilla2021FashionSketches,Chowdhury2022GarmentModeling, liu2024automatic}.
Recent studies~\cite{liu2023towards,he2024dresscode,korosteleva2022neuraltailor,korosteleva2023garmentcode,bian2024chatgarment,nakayama2024aipparel,zhou2024design2garmentcode} on sewing patterns start to focus on autonomously generating diverse sewing patterns through different conditions rather than merely adjusting or producing a single garment.
One of the recent SOTA methods, DressCode~\cite{he2024dresscode}, first generates garments through natural language and yields visual-appealing appearances.
However, the capability of DressCode is limited in modeling complex sewing patterns, and furthermore, it does not consider the relation between garments and body shapes, difficult to drape on various bodies.
Another typical work, parametric sewing pattern~\cite{korosteleva2023garmentcode}, can control complex sewing patterns, while it requires predefined templates and a delicate selection of different control scale values, which is not user-friendly.
Different from these methods, our SewingLDM not only has the ability to represent complex sewing patterns, but can also generate sewing patterns based on multi-modal intuitive conditions, \ie, natural language, garment sketches, and body shape.
These capabilities enable easily creating tailored garments that conform precisely to individual body shapes.

\section{Method}

\begin{figure}[t]
  \centering
  \includegraphics[width=\linewidth]{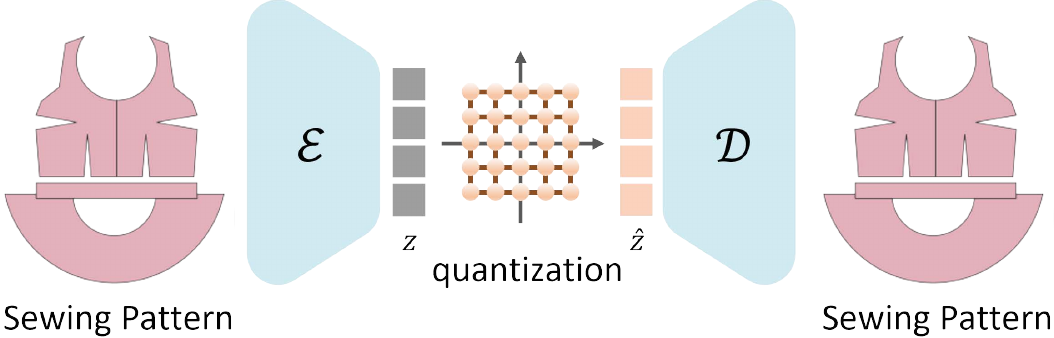}
   \caption{\textbf{Sewing pattern compression.} We compress the sewing pattern representations into a bound and compact latent space.}
   \vspace{-1.5em}
   \label{fig:compression}
\end{figure}

\begin{figure*}[t]
  \centering
  \includegraphics[width=\linewidth]{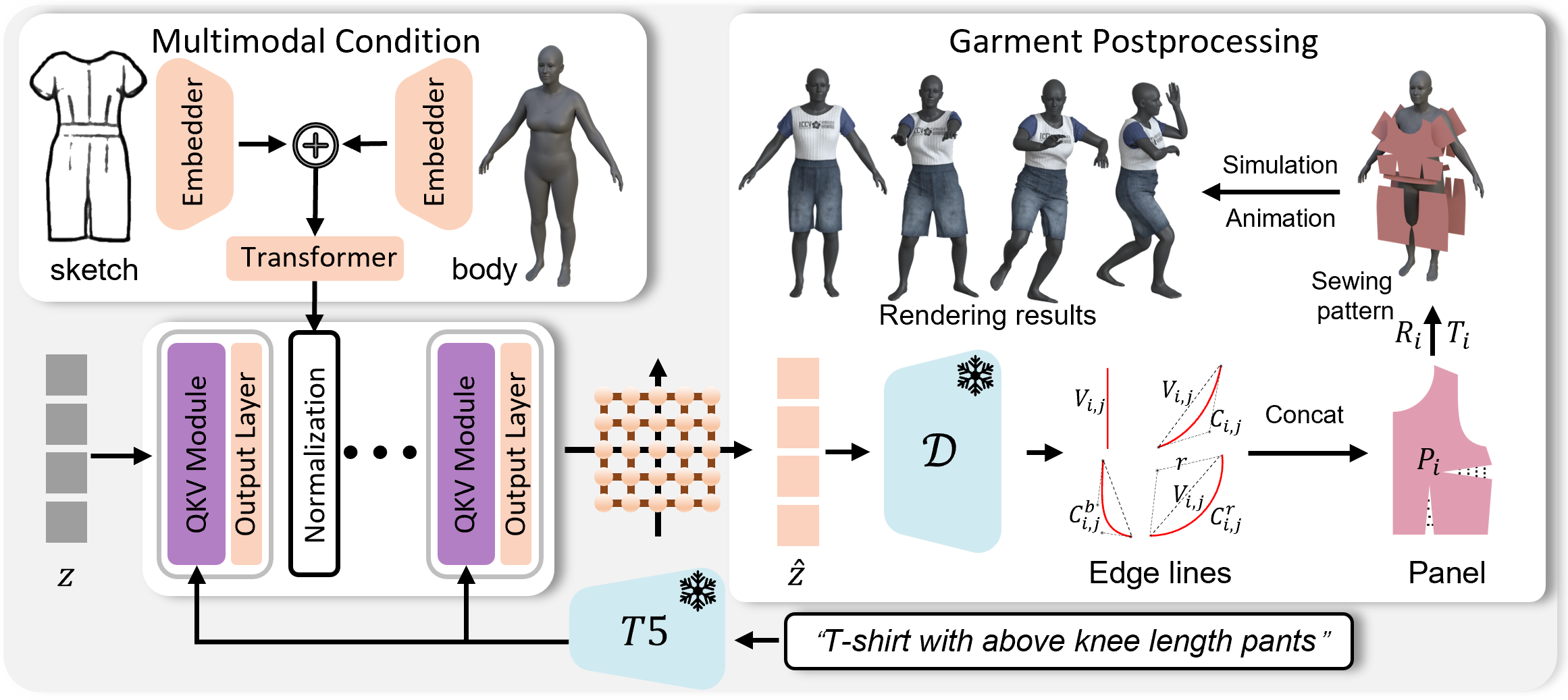}
   \caption{\textbf{Multimodal latent diffusion model.} 
   After training the text-guided diffusion model, we fuse the features of sketches and body shapes and normalize them into the diffusion model, with minimal fine-tuning parameters of the diffusion model. The trained network parameters are depicted in orange, while the frozen parameters are shown in purple. The output latent is then quantized into a designed latent space and serves as the input of the decoder to yield all edge lines. Edge lines connect from beginning to end to form panels, placed on the corresponding body regions. Finally, we can get suited garments through the modern CG pipeline.}
   \vspace{-1.5em}
   \label{fig:pipeline}
\end{figure*}

To generate garments suited for various humans, we introduce \textbf{SewingLDM}, a latent-based diffusion model, to create complex 3D sewing patterns, conditioned by personalized body shapes, text prompts, and garment sketches.
We first review the original sewing pattern representation~\cite{korosteleva2022neuraltailor} (\cref{sec:preliminaries}) and then we improve it with special designs to cover complex sewing patterns (\cref{sec:representation}), as illustrated in~\cref{fig:sewing_pattern_pre}.
Subsequently, we compress the sewing pattern representation into a compact latent space enabling the generation of complex sewing patterns with centimeter-level precision in~\cref{fig:compression} (\cref{sec:compression}).
Finally, we train a latent diffusion model under multi-modal conditions through a two-stage training strategy, as illustrated in~\cref{fig:pipeline} (\cref{sec:generation}).
Based on the proposed framework, SewingLDM can generate complex garments based on the body shape and align with the user-provided text description or garment sketches.

\subsection{Preliminaries on Sewing Pattern}
\label{sec:preliminaries}
Sewing patterns are CAD representations of garments, representing 2D shapes and 3D placement of cloth, as shown in~\cref{fig:sewing_pattern_pre}.
NeuralTailor~\cite{korosteleva2022neuraltailor} first transfers sewing patterns into vector representations as inputs of the neural network.
The sewing pattern contains $N_p$ panels $\{P_i\}_{i=1}^{N_p}$, with each panel $P_i$ including $N_i$ edges $\{E_{i,j}\}_{j=1}^{N_i}$.
For each edge $E_{i,j}$, a vector $\{V_{i,j}\}_{j=1}^{N_i}\in\mathbb{R}^2$ is utilized to represent the direction from its starting to ending point.
In NeuralTailor~\cite{korosteleva2022neuraltailor}, sewing patterns only have two kinds of edges, \ie, straight lines and quadratic lines.
Quadratic lines use two additional parameters $C_{i,j} = (c_x, c_y)$ representing the control point of the Bezier curve.
Rotation $R_i\in SO(3)$ and translation $T_i \in\mathbb{R}^3$ are utilized to represent the 3D placement of each panel $P_i$.
Moreover, to depict the stitching connecting each inner or outer panel edge, it incorporates per-edge stitch tags $\{S_{i,j}\in\mathbb{R}^3\}_{j=1}^{N_i}$ and stitch masks $\{M_{i,j}\in\{0,1\}\}_{j=1}^{N_i}$.
The stitch tag $S_{i,j}$ is determined by the 3D position of the corresponding edge, which utilizes the Euclidean distance between edges as a measure of stitch similarity.
The stitch mask $M_{i,j}$ is a binary flag to indicate whether there are stitches on the edge.

\subsection{Extended Representation for Sewing Pattern}
\label{sec:representation}

For more complex garment designs, the modern industry will use special designs to make garments more fashionable, as illustrated in~\cref{fig:sewing_pattern_pre}.
Original sewing pattern representation in~\cite{korosteleva2022neuraltailor} can not cover complex garments with more kinds of curve lines, \ie, cubic and circle lines, and additional attachment constraints in collars and waistbands to prevent cloth sliding from human bodies for certain garments, \eg, strapless tops, and loose pants.
Moreover, the stitches may intersect for some edge pairs, causing errors during simulation.
To precisely represent complex clothes, we extend each edge feature to high dimensions to cover complex patterns.
Then we preprocess them into a uniform tensor shape to feed into the neural network.

\noindent \textbf{Representation.}
For each edge $E_{i,j}$, we append origin control parameters $C_{i,j}$ with cubic line control parameters $C^{b}_{i,j}\in\mathbb{R}^4$, representing two control points, and circle line control parameters $C^r_{i,j}\in\mathbb{R}^3$, which represents the radius $r$ and the rotation angle, to cover more kinds of curve lines.
Further, we use two binary flags $\{E^t_{i,j,k}\in\{0,1\}\}_{k=1}^{2}$ to denote these 4 different edge types.
Moreover, 3 specific binary flags $\{A_{i,j, k}\in\{0,1\}\}_{k=1}^{3}$ are included to indicate the attachment type for certain edges, such as those associated with the collar and waistband, to prevent garments from sliding during simulation.
We add one binary flag to ${M_{i,j}}$ as the new stitch mask $\{M_{i,j,k}^{'}\in\{0,1\}\}_{k=1}^{2}$ to additionally denote whether the stitch direction needs to be reversed to prevent stitch intersections.

\noindent \textbf{Preprocessing.}
Before input to the neural networks, the vector representation needs to be the same size for all data during training.
For each edge $E_{i,j}$, we concatenate all extended parameters and append with rotation $R_i$ and translation $T_i$ of panel $P_i$ to form the high-dimensional edge feature $E^f_{i,j}$.
Furthermore, we design a binary flag $\{E^m_{i,j}\in\{0,1\}\}_{j=1}^{N_i}$ to denote the existence of each edge.
All features are concatenated to form a 29-dimensional vector for each edge feature $E^f_{i,j}$, represented as follows:
\begin{equation}
    \begin{aligned}
        E^f_{i,j} = V_{i,j} & \oplus C_{i,j}\oplus C^b_{i,j} \oplus C^r_{i,j} \oplus S_{i,j} \oplus R_i \\
        &\oplus T_i \oplus E^t_{i,j} \oplus E^m_{i,j} \oplus A_{i,j}\oplus M_{i,j}^{'},
    \end{aligned}
\end{equation}
where $i$ is in the range of $[1,max(N_p)]$, $j$ is in the range of $[1,max(N_i)]$.

Then, all edge features $\{E^f_{i,j}\}_{j=1}^{N_i}$ are concatenated and padded with 0 to max edge number and max panel number to get the representation of sewing pattern $F$, in the shape of $(max(N_p)\times max(N_i), 29)$.
Before input to the neural network, all continuous values are standardized, and all binary flags are transformed into $\{-1, 1\}$.

\subsection{Compact Latent for Sewing Pattern}
\label{sec:compression}

The vector representation $F$ inevitably incorporates redundant information as the panel and edge numbers increase, preventing generation models from learning the distribution of $F$.
Using the tokenization in the previous method~\cite{he2024dresscode} will lead to excessive GPU consumption, exceeding hardware limitations.
As indicated by the recent compression methods~\cite{zhao2024image, DBLP:conf/iclr/MentzerMAT24, DBLP:conf/iclr/YuLGVSMCGGHG0ER24}, it is necessary to compress $F$ into a compact latent space and maintain the reconstruction quality.
Following this idea, we train an auto-encoder to compress and quantize the $F$ into a latent space where each dimension is bounded within the range $\left[-1, 1\right]$, as illustrated in~\cref{fig:compression}.
The sewing pattern representation $F$ is encoded to $z$ by the encoder $\mathcal{E}$ and quantized to $\hat{z}$ in the constrained latent space, subsequently reconstructed by the decoder $\mathcal{D}$. The process can be represented as:
\begin{equation}
    z = \mathcal{E}(F_{\mathrm{gt}}), \\
    \hat{z} = \frac{round(n \times tanh(z))}{n}, \\
    F_{\mathrm{rec}} = \mathcal{D}(\hat{z}),
    \label{eq:auto-encoder}
\end{equation}
where $n$ is an integer used to modify the spacing between each $\hat{z}$ in the latent space.
For training the encoder $\mathcal{E}$ and decoder $\mathcal{D}$, we combine the loss in previous works~\cite{korosteleva2022neuraltailor, DBLP:conf/iclr/MentzerMAT24} with additional binary cross-entropy loss $\mathcal{L}_{\mathrm{BCE}}$ to constrain the newly incorporated binary flags:
\begin{equation}
    \mathcal{L}_{\mathrm{total}} = \lambda_1\mathcal{L}_{\mathrm{rec}} + \lambda_2\mathcal{L}_{\mathrm{panel}} + \lambda_3\mathcal{L}_{\mathrm{stitch}} + \lambda_4\mathcal{L}_{\mathrm{BCE}},
\end{equation}
where $\lambda_1,\lambda_2,\lambda_3,\lambda_4$ are hyperparameters to balance each loss term.
$\mathcal{L}_{\mathrm{rec}}$ is the MSE loss to keep reconstruction quality, while $\mathcal{L}_{\mathrm{panel}}$ and $\mathcal{L}_{\mathrm{stitch}}$ proposed in ~\cite{korosteleva2022neuraltailor} to ensure the integrity of the garments.

After training, the sewing pattern representation $F$ can be efficiently compressed into a bounded and compact latent space without compromising important information.
Moreover, to facilitate the learning of generation models, each dimension of the latent is evenly distributed within the coordinates $\{-1, -0.5, 0, 0.5, 1\}$ by setting $n=2$ in~\cref{eq:auto-encoder}.

\subsection{Multimodal Conditions of Diffusion Model}
\label{sec:generation}
Inspired by the great power of controlled generation in the diffusion model, we employ latent diffusion~\cite{rombach2022high} as our generation model.
Our generation model is based on the DiT architecture~\cite{peebles2023scalable,chen2024pixart}, which is scalable to different sizes of sewing patterns.
To balance multi-modal conditions and facilitate future conditional scalability, we design a two-step training strategy: 1) In the first step, we train the latent diffusion model with IDDPM loss~\cite{nichol2021improved} only under the text guidance extracted by T5 tokenizer~\cite{raffel2020exploring}; 2) In the second step, we embed the knowledge of body shapes and garment sketches into the diffusion model for detailed control and body-suited garment generation, as depicted in~\cref{fig:pipeline}.

The text-guided latent diffusion model serves as a fundamental model for extensive multi-modal conditions injection.
For the injection of body shapes and garment sketches, a naive idea is to inject them through two ControlNet~\cite{zhang2023adding} branches.
While sketches will change along with body shape, \eg, sketches will get wider when bodies grow fatter.
We propose to first use embedders to extract the feature from sketches and body shapes, and then simply concatenate them together, and input them into a light transformer layer.
During the light transformer layer, the features of sketches and body shapes can be thoroughly fused and get the relation between each other through self-attention modules and output as $\boldsymbol{F}_{bs}$.
Then we normalize mean $\boldsymbol{\mu}_{bs}$ and variance $\boldsymbol{\sigma}_{bs}$ of $\boldsymbol{F}_{bs}$ into the same mean $\boldsymbol{\mu}_z$ and variance $\boldsymbol{\sigma}_z$ with the latent features $\boldsymbol{F}_z$, serving as a residual to control the output.
The new $\hat{\boldsymbol{F}}_z$ is represented as below:
\begin{equation}
    \hat{\boldsymbol{F}}_z=\frac{(\boldsymbol{F}_{bs}-\boldsymbol{\mu}_{bs}) \times \boldsymbol{\sigma}_{bs}}{\boldsymbol{\sigma}_z+\epsilon} + \boldsymbol{\mu}_z + \boldsymbol{F}_z,
\end{equation}
where $\epsilon$ is a small constant for numerical stability.
After the normalization, we assume the newly added residual is similar to $\boldsymbol{F}_z$, which does not need to retrain the whole diffusion model in the second stage. 
We only fine-tune the output layer of the attention modules in each DiT block to transform the normalized features into the desired distribution.
After two-stage training, our generation model can precisely follow the text guidance and sketches under various human shapes, enabling more body-suited and detailed controlled garment generation for individuals.

\section{Experiments}

\subsection{Experiment Setup}
\noindent\textbf{Dataset.} To train a generation model under the condition of texts or sketches, it is essential to acquire the corresponding paired data.
We extend the current dataset~\cite{korosteleva2024garmentcodedata} with additional textual annotations and garment sketches.
The dataset~\cite{korosteleva2024garmentcodedata} consists of 120,000 sewing patterns, covering a variety of clothing styles for different body types.
Sewing patterns in~\cite{korosteleva2024garmentcodedata} consider the relationship between various body shapes and garments, resulting in garments that are well-tailored to individual body types.
Building on~\cite{korosteleva2024garmentcodedata}, we annotate each garment with text prompts according to its design parameters file and refine it with GPT-4~\cite{achiam2023gpt}, resulting in detailed text annotations.
However, relying solely on textual descriptions may not precisely dictate garment shapes, potentially yielding undesirable outputs.
To enhance control over the generation, we propose to generate richer annotations like sketches.
For each garment, we utilize PiDiNet~\cite{pdc-PAMI2023}, a pre-trained edge detection network, to extract garment sketches, thereby enriching the design details of the garment.
More annotation details are represented in the supplementary material.

\noindent\textbf{Implementation Details.} We train our model on 4 RTX A6000 GPUs with 48G memory, where the auto-encoder requires 12 hours for training.
The hyperparameters for training the auto-encoder, $\lambda_1,\lambda_2,\lambda_3,\lambda_4$ are set as $5, 1, 1, 1$.
Training the text-guided latent diffusion model takes 2 days in the first stage, and training the multi-modal conditions requires an additional 10 hours to reach convergence in the second stage.
During the second stage, the sketch or text conditions are set to zero with a probability of 0.25 to ensure the model retains the capability to generate desired garments based on a single input condition.

\begin{figure}[t]
  \centering
  \includegraphics[width=\linewidth]{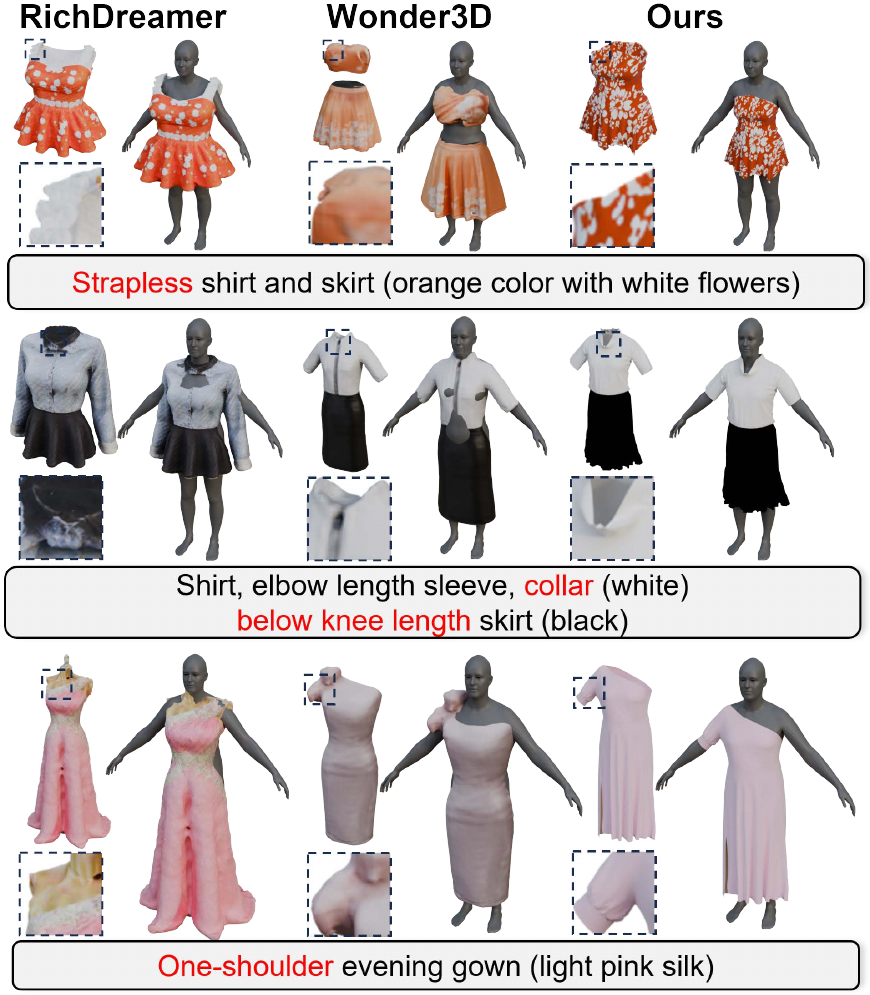}
  \vspace{-1.5em}
   \caption{\textbf{Comparison with 3D mesh generation method.} We present the garments and draping results for each method. Our method successfully generates modern design garments with remarkable visual quality and close fitting to various body shapes. In contrast, Wonder3D~\cite{long2024wonder3d} and RichDreamer~\cite{qiu2024richdreamer} only generate close-surface meshes and contain obvious artifacts, resulting in human bodies clipping through the garments.}
   \vspace{-1.5em}
   \label{fig:3D-generation}
\end{figure}

\subsection{Qualitative Comparison}
We conduct qualitative comparisons with SOTA mesh generation methods and sewing pattern generation methods, respectively, to demonstrate our CG-friendly and superior generation results for various body shapes.

\begin{figure*}[t]
  \centering
  \includegraphics[width=\linewidth]{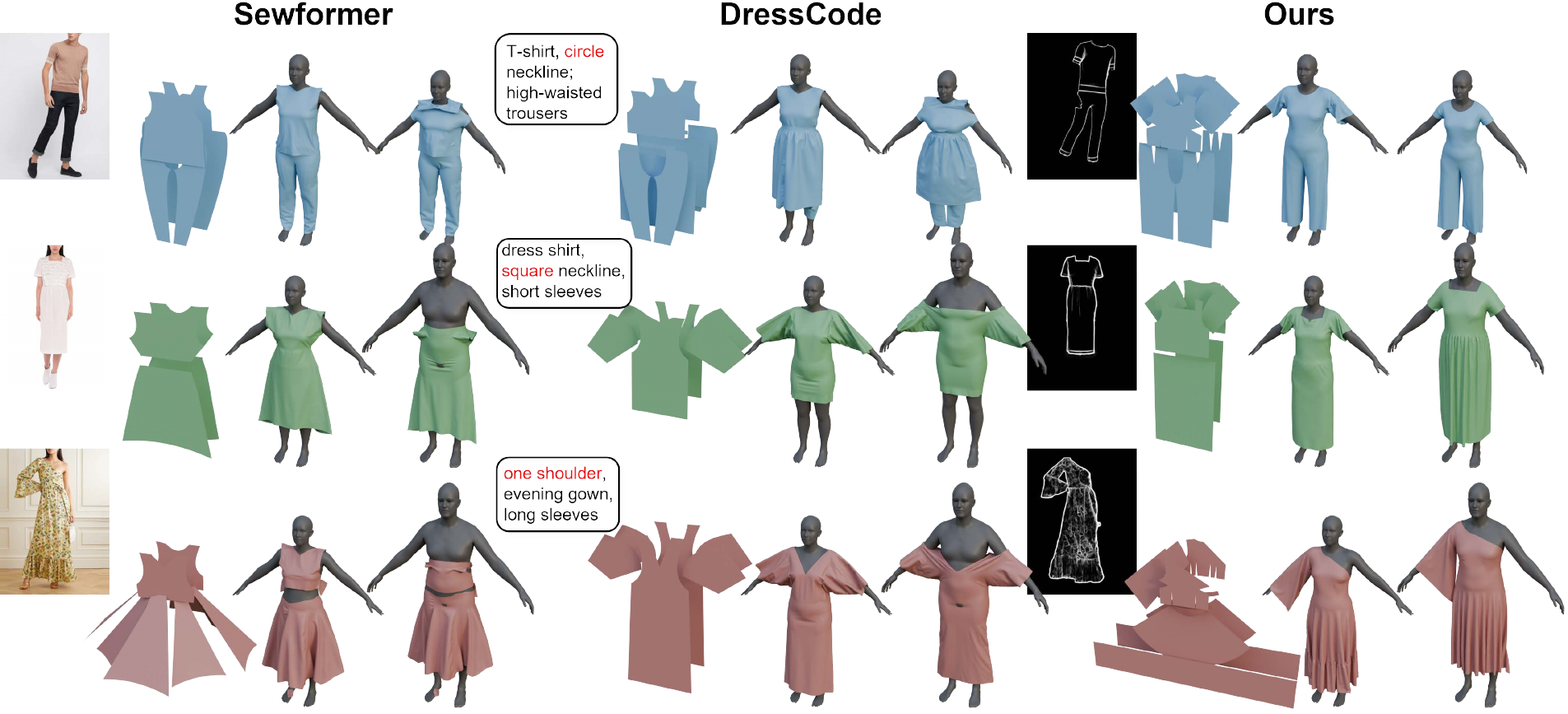}
   \vspace{-1.5em}
   \caption{\textbf{Comparison of sewing pattern generation for various body shapes.} For each method, we present corresponding conditions and generated results, including sewing patterns, draping results on average body shape, and draping results on another body shape. Our method can generate complicated sewing patterns aligned with sketch conditions and text prompts, draping on various body shapes. In contrast, Sewformer~\cite{liu2023towards} and DressCode~\cite{he2024dresscode} both fail to generate complex and body-suited garments.
   }
    \vspace{-0.5em}
    \label{fig:compare_sewing_pattern}
\end{figure*}

\noindent\textbf{Comparison on 3D Mesh Generation Methods.} We compare our SewingLDM with the current SOTA 3D mesh generation methods, \ie, Wonder3D~\cite{long2024wonder3d} and RichDreamer~\cite{qiu2024richdreamer} with the same text prompts containing both the geometry and texture information.
Note that, RichDreamer~\cite{qiu2024richdreamer} is an image-guided generation method; thus, we feed the sketches and text prompts to ControlNet~\cite{zhang2023adding} with Stable Diffusion~\cite{rombach2022high} to generate corresponding images.
As the qualitative comparisons shown in~\cref{fig:3D-generation}, Wonder3D and RichDreamer can both generate garments aligned with text prompts.
However, both of them are close-surface meshes and do not consider human body shapes, resulting in obvious clipping when draping on human bodies.
In contrast, our method generates sewing patterns for various human bodies through two-stage training, which are easy to drape on human bodies and maintain the physical properties of clothing with fantastic clothes wrinkles.
The results show that our garments are all well-fitted with complex geometry and aligned with the conditions.

\noindent\textbf{Comparison of Sewing Pattern Generation Methods.} We also compare our method with current SOTA sewing pattern generation methods, \ie, DressCode~\cite{he2024dresscode} and Sewformer~\cite{liu2023towards}.
To ensure a fair comparison, we directly utilize the in-the-wild images chosen from the Multimodal Garment Designer~\cite{baldrati2023multimodal} and transfer into the corresponding conditions of DressCode and Sewformer.
Moreover, since Sewformer and DressCode are only trained with the \textbf{average} SMPL body shapes, we drape the generative sewing patterns onto the average body shape and another body shape to validate the effectiveness of body-aware garment generation. 
As illustrated in~\cref{fig:compare_sewing_pattern}, both Sewformer and DressCode fail to generate textual-aligned garments due to the complex garment descriptions.
Moreover, since these baseline methods do not account for various bodies and generate attachments to any edge of sewing patterns, they can not be worn on diverse body shapes, only on the average body shape, sliding from the body or just failing to simulate the results.
In contrast, our method uses a compact latent space to represent complex sewing patterns with centimeter-level precision, which greatly improves the generation ability on complex sewing pattern generation, \eg, fitted shirts, one-shoulder gowns, and square necklines.
Moreover, due to the body shape conditions, our method can fit various body shapes, which provides a more user-friendly approach to getting the desired made-to-measure garments.
The results demonstrate the superiority of our method in complex garment design and various body adaptability.
We also provide concurrent work comparison and more in-the-wild and in-domain conditions in the supplementary material.

\begin{table}[t]
\begin{center}
\scriptsize
\renewcommand\arraystretch{1.5}
\setlength{\tabcolsep}{.3em}
\resizebox{\linewidth}{!}{
\begin{tabular}{c ccccc}
\toprule
 & RichDreamer & Wonder3D & Sewformer & Dresscode & Ours \\
\hline
Runtime $\downarrow$ & $\sim$ 4 mins & $\sim$ 4 hours & \textbf{$\sim$ 3 mins} & \textbf{$\sim$ 3 mins} & \textbf{$\sim$ 3 mins} \\

\makecell[c]{Clothes-to-body \\ Distance} $\downarrow$ & 6.19 cm & 6.54 cm  & 5.45 cm  & 3.69 cm & \textbf{2.20 cm} \\

Users Values $\uparrow$ & 1.89 & 1.88 & 2.10 & 3.56 & \textbf{4.60} \\
\bottomrule
\end{tabular}
}
\end{center}
\vspace{-1.5em}
\caption{\textbf{Quantitative Comparison.} We compare the generation efficiency, the average clothes-to-body distance, and users' evaluation. All metrics show that our method generated superior results.}
\vspace{-1.5em}
\label{tab:quantitative}
\end{table}

\begin{table*}[t]
    \centering
    \scriptsize
    \begin{tabular}{lccccccc}
    \toprule
    Method &  Panel L2$(\downarrow)$ & Panel Acc$(\uparrow)$ & Edge Acc$(\uparrow)$  & Rot L2$(\downarrow)$ & Transl L2$(\downarrow)$ & Stitch Acc$(\uparrow)$ & Failure Rate$(\downarrow)$ \\\midrule
    SewFormer{*} & 12.3 & 79.4 & 44.7 & .0400 & 4.5 & 2.8 & 4.3\% \\
    \textbf{AE~(Ours)} & \textbf{0.64} & \textbf{99.8} & \textbf{88.5} & \textbf{.0004} & \textbf{0.6} &  \textbf{90.8} & \textbf{0}\\
    \textbf{SewingLDM}   & \textbf{3.13} & \textbf{97.8} & \textbf{82.7} & \textbf{.0043} & \textbf{1.2} & \textbf{84.2} & \textbf{0} \\
    \bottomrule
    \end{tabular}
    \caption{
    \textbf{Reconstruction metrics.} 
    We report the reconstruction performance on the test set of the GarmentCode dataset~\cite{korosteleva2024garmentcodedata}.
    Our auto-encoder (AE) can reconstruct the sewing pattern with centimeter-level precision compliant with industrial standards.
    }
    \label{tab:metrics}
\end{table*}

\subsection{Quantitative Comparison}

Besides qualitative comparisons, we also perform quantitative comparisons with these SOTA methods, evaluating aspects including generation efficiency, clothes-to-body distance, and user study.
The clothes-to-body distance is an essential metric for body-aware garment generation, indicating whether the clothes are close-fitted to body shapes calculated by averaging the minimum distance from each garment point to human bodies.
Besides, we also perform a user study to further assess the quality of garment generation.  
We take 10 text prompts to generate diverse garments and render the generated results, draping on different body shapes for each method. 
Then we ask 30 users to give a value of these rendering results with comprehensive consideration for two aspects: 1) consistency with text descriptions; and 2) well-fitting with human bodies. 
As illustrated in~\cref{tab:quantitative}, the preference results demonstrate a notable superiority of our method over SOTA approaches in both aspects, highlighting in generating garments that are both well-suited to various bodies and exhibit high fidelity aligned with text descriptions.
Furthermore, we also evaluate sewing pattern reconstruction metrics as proposed in SewFormer~\cite{liu2023towards}, including: 1) Panel L2, Rot L2, and Transl L2 (average edge, rotation, and translation L2 distances between predicted and ground-truth patterns); 2) Panel Accuracy, Edge Accuracy, Stitch Accuracy (accuracy of panel count, edge prediction per panel, stitch prediction); and 3) Failure Rate (simulation failure percentage).
All L2 metrics are in centimeters, except rotation in radians.
To ensure a fair evaluation, SewFormer{*} is fine-tuned on the GarmentCode dataset until its validation loss no longer improves.
Dresscode, which requires over 30k tokens for complex patterns and exceeds current GPU limitations, unabling to train on the GarmentCode dataset, was evaluated only for failure rate (11.4\%) using its open-source code and checkpoint.
As shown in~\cref{tab:metrics}, our method achieved superior performance compared with baselines on complex sewing pattern generation as our latent space is well-designed and suitable for centimeter-level garments compression and our model takes the body shapes into account.

\subsection{Ablation Study}
\begin{figure*}[t]
  \centering
  \includegraphics[width=\linewidth]{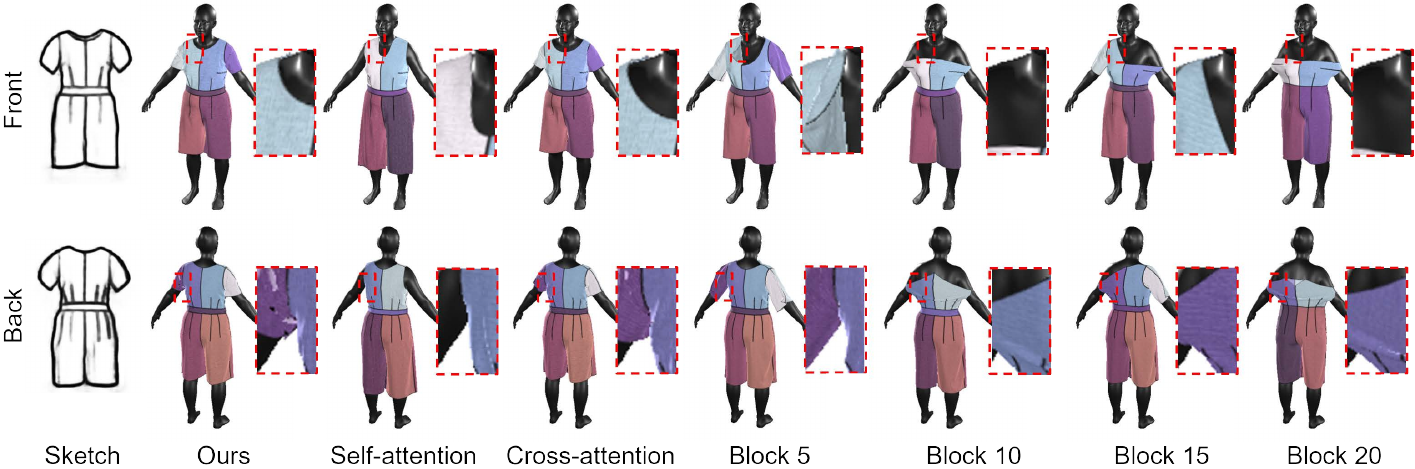}
  \vspace{-1.5em}
   \caption{\textbf{Ablation of the multi-modal condition.} We have taken an ablation experiment on the training parameters of the output layer in different attention modules. We also explore the relationship between results and injection positions.}
   \label{fig:params_ablation}
   \vspace{-1.5em}
\end{figure*}

\noindent\textbf{Sewing Pattern Compression.} To maintain the reconstruction ability and dense compression of sewing patterns in the meantime, we have explored numerous parameters for the compression network, as shown in~\cref{tab:compression}.
We measure the reconstruction ability, generation ability, clothes-to-body distance, and codebook usage under different settings.
The codebook usage is calculated by the number of used latent $N_U$ dividing the latent number in latent space $N_L$ as follows:
\begin{equation}
    codebook~usage = \frac{N_U}{N_L} = \frac{N_U}{{(2n+1)}^{n_f}},
\end{equation}
where $n$ is an integer in pre-defined~\cref{eq:auto-encoder} for quantization, $n_f$ is the last dimension length of latent.
As illustrated in~\cref{tab:compression}, without compression or lower compression, the latent space is inappropriate for the generation model to learn the distribution of latent.
In contrast, with a compact latent space, the latent is fully utilized, resulting in a well-generation ability and various body adaptability.

\begin{table}[t]
\begin{center}
\scriptsize
\renewcommand\arraystretch{1.2}
\setlength{\tabcolsep}{.3em}
\resizebox{\linewidth}{!}{
\begin{tabular}{c cccccc}
\toprule
\textbf{Compression shape} & \makecell[c]{w/o \\ compression} & \makecell[c]{256*32 \\ n=32} & \makecell[c]{256*12 \\ n=8} & \makecell[c]{256*8 \\ n=2} & \makecell[c]{256*6 \\ n=2} & \makecell[c]{256*4 \\ n=2} \\
\hline
Reconstruction & \checkmark  & \checkmark  & \checkmark  & \checkmark & \checkmark & $\times$ \\

Generation & $\times$  & $\times$  & $\times$  & \checkmark & \checkmark & - \\

\makecell[c]{Clothes-to-body \\ Distance} $\downarrow$ & - & -  & -  & 2.87 cm & \textbf{2.20 cm} & - \\

Codebook usage & - & 0\%  & 0\%  & 91\% & 100\% & - \\
\bottomrule
\end{tabular}
}
\end{center}
\vspace{-1.5em}
\caption{\textbf{Different Compression.} We try different compression shapes for the latent and set the different values of n. \checkmark means it can well do this task, while $\times$ means it fails in this task.}
\vspace{-1.5em}
\label{tab:compression}
\end{table}

\noindent\textbf{Multi-modal Controllable Generation.} In the context of injecting body shape and garment sketch conditions, we perform ablation studies on the optimized parameters of the output layers across different attention modules, \ie, both self-attention and cross-attention, self-attention only, and cross-attention only.
By default, we inject the additional condition after the first transformer block.
Moreover, we investigate the impact of different injection positions, specifically after block 5, block 10, block 15, and block 20, as illustrated in ~\cref{fig:params_ablation}.
Notably, optimizing in both attention results in more desired circle necklines than only optimizing in cross-attention and is better aligned with sketches compared with only training output layers in self-attention, which fails to generate the desired neckline and sleeves.
This shows that the additional conditional can closely resemble the latent features, which needs more learning across text prompts and latents rather than within latents alone.
Consequently, during the ablation study on injection positions, we fine-tune the output layers of both attention for better results.
We observe that as the layer depth increases, the garment gradually loses key components, \eg, sleeves or waistband, resulting in unable draping on the human body.
In contrast, injecting the condition at shallower layers facilitates better fusion of the additional condition with the combined latent feature, leading to more accurate results.
In summary, we train the output layers of both attention and inject the additional condition after block 0, which yields optimal results.
\section{Conclusion and Limitations}
\begin{figure}[t]
  \centering
  \includegraphics[width=\linewidth]{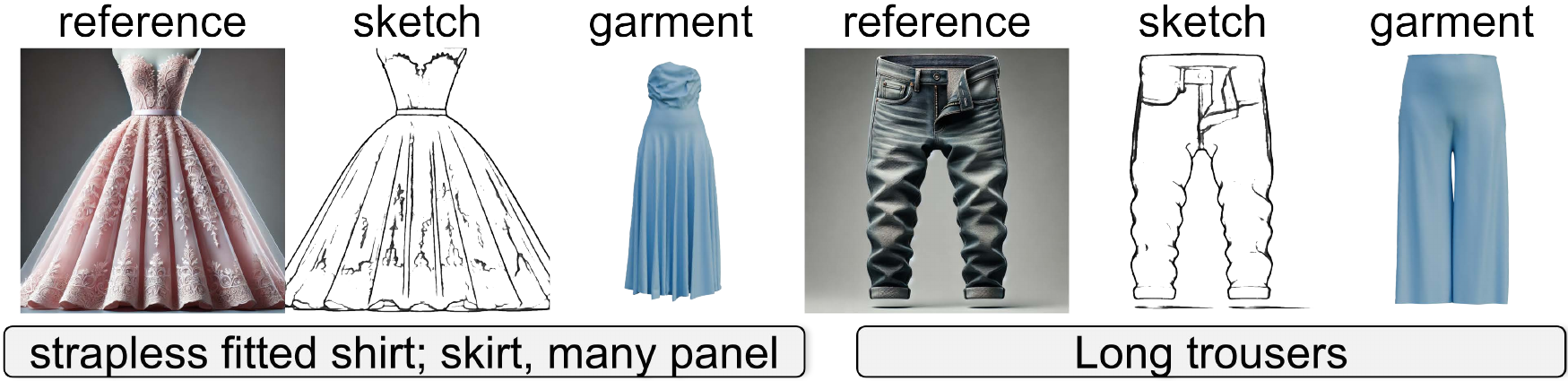}
   \caption{\textbf{Limitations.} For intricate sketches, such as bridal gowns or additional accessories like pockets and zippers, our method may fail to generate the desired garments.}
   \label{fig:limitations}
   \vspace{-1.5em}
\end{figure}

In conclusion, our SewingLDM can generate complex sewing patterns under the condition of text prompts, garment sketches, and body shapes.
We propose an enhanced vector representation of sewing patterns and compress them into a bounded and compact latent space for more generalized garment designs and facilitating the training of the diffusion model.
To accommodate multi-modal conditioning and future conditions, we introduce a two-step training strategy.
We first train a latent diffusion model only conditioned by text prompts.
Subsequently, we incorporate the condition of garment sketches and body shapes by optimizing the output layers of the attention modules while maintaining the responsiveness to text-based guidance.
Finally, our generation model can be conditioned by multi-modal input, resulting in body-suited generation and detailed control of garments.

Despite the promising results, our method still has several limitations that should be addressed in future work, as illustrated in~\cref{fig:limitations}.
The major limitation is that our method encounters challenges with certain modern designs, \eg, zippers, and pockets. 
Another limitation is that it occasionally struggles with aligning complex sketches of intricate garments.
Our further work aims to explore comprehensive representations of daily garments and expand the range of conditions applicable during the generation process.

{
    \small
    \bibliographystyle{ieeenat_fullname}
    \bibliography{main}
}

\clearpage
\setcounter{page}{1}
\maketitlesupplementary
 
In this supplementary document, we first provide comprehensive details of complex sewing patterns (\cref{sec:details}). Afterward, we compare with the parametric method~\cite{korosteleva2023garmentcode}, which needs a delicate selection of values and professional knowledge of garment designs (\cref{sec:parametric}).
We also compare with concurrent open-sourced methods (\cref{sec:concurrent}).
Then, we provide the annotation details of our dataset to better demonstrate the precision and accuracy of the dataset (\cref{sec:annotation}).
To substantiate the efficacy of extended representation and our two-step training strategy, we perform an ablation study on representation (\cref{sec:extend_rep}) and the training strategy (\cref{sec:ablation}).
Additionally, we provide examples of our user study, which ensures a fair and objective evaluation of our method compared to others (\cref{sec:user_study}).
We further include a user case (\cref{sec:use_case}) and examples of generated garments that demonstrate the robustness and generative capabilities of our method across various sketches and body types (\cref{sec:qualitave}).

\section{Representation Details}
\label{sec:details}
\begin{figure}[t]
  \centering
  \includegraphics[width=\linewidth]{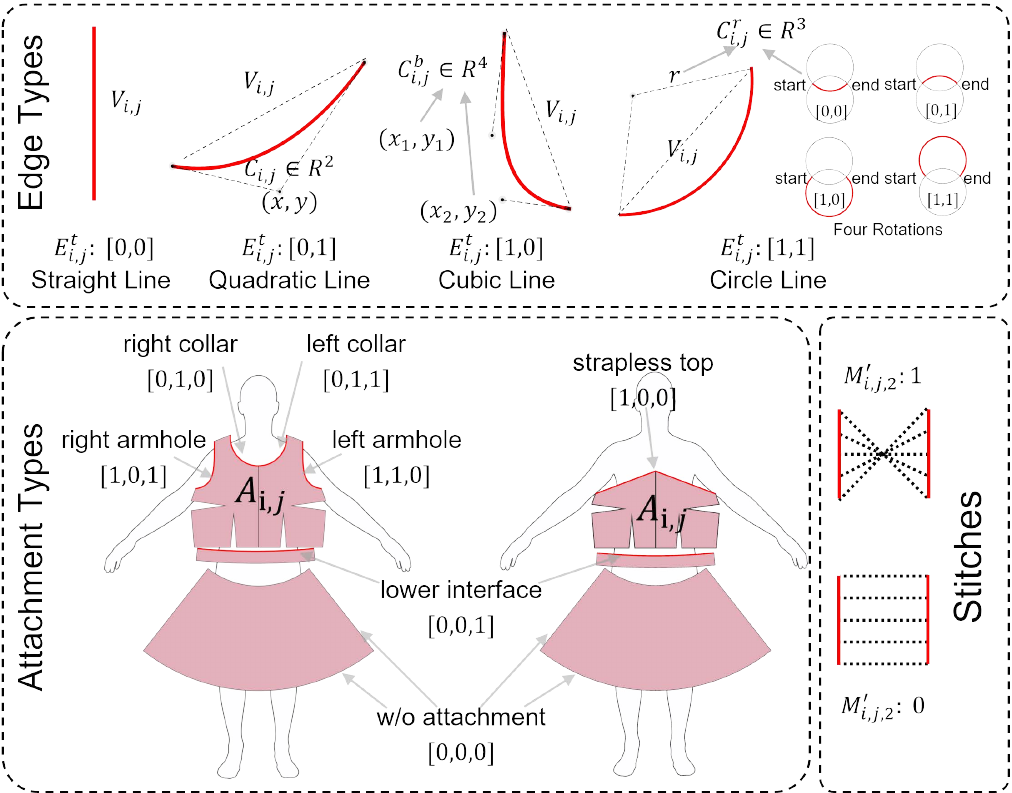}
  \vspace{-1.5em}
   \caption{\textbf{Representation details.} We present various kinds of edges, attachments, and stitches with detailed annotations.}
   \vspace{-1.5em}
   \label{fig:representation_details}
\end{figure}

\textbf{Extended Representation.} The binary concrete representations of different edge types, attachment types, and stitches are depicted in \cref{fig:representation_details} alongside their corresponding annotations.
For edges, in addition to the vector $V_{i,j}$ representing from the start point to the endpoint, the cubic line employs the control parameters $C^{b}_{i,j}\in\mathbb{R}^4$ to define two control points $(x_1,y_1)$ and $(x_2,y_2)$ in the 2D coordinate.
The circle line uses additional control parameters $C^r_{i,j}\in\mathbb{R}^3$, which specify the radius $r$ and four rotations with two binary flags, including the counterclockwise acute angle ($[0, 0]$), the clockwise acute angle ($[0, 1]$), the counterclockwise reflex angle ($[1, 0]$), and the clockwise reflex angle ($[1, 1]$).
Furthermore, edge types are denoted as follows: $E^t_{i,j} = [0, 0]$ for the straight line, $E^t_{i,j} = [0, 1]$ for the quadratic line, $E^t_{i,j} = [0, 0]$ for the cubic line, and $E^t_{i,j} = [1, 1]$ for the circle line.
The attachments are visually distinguished by highlighting the associated edges in red and annotating them with the name and value of $A_{i,j}$.
Edges without attachment are not highlighted and use the default value of $A_{i,j}= [0,0,0]$.
There are six kinds of attachment types, \ie, lower interface ($[0, 0, 1]$), right collar ($[0,1,0]$), left collar ($[0,1,1]$), strapless top ($[1, 0, 0]$), right armhole ($[1,1,0]$), and left armhole ($[1,0,1]$).
For reversal stitch $\{M_{i,j,2}^{'}\in\{0,1\}\}$, 0 means the stitch direction does not need reversal, while 1 means the stitch direction needs to be reversed.
With the detailed representation of sewing patterns, users can convert the sewing patterns into vector representations as input for neural networks.

\noindent\textbf{Panel Order invariance.}
Since all panels in the dataset are fixed, we define an order from the upper body to the lower body to construct the vector representation, guaranteeing the panel order invariance.
The panel order is represented as below:
\begin{lstlisting}[language=json,firstnumber=1,basicstyle=\ttfamily\footnotesize]
{
    "1": ["right_sleeve_b"],
    "2": ["right_sleeve_f"],
    "3": ["right_ftorso"],
    "4": ["right_btorso"],
    "5": ["left_ftorso"],
    "6": ["left_btorso"],
    "7": ["left_sleeve_f"],
    "8": ["left_sleeve_b"],
    "9": ["left_collar_back"],
    "10": ["left_collar_front"],
    "11": ["right_collar_front"],
    "12": ["right_collar_back"],
    "13": ["sl_right_cuff_f"],
    "14": ["sl_right_cuff_b"],
    "15": ["sl_right_cuff_skirt_f"],
    "16": ["sl_right_cuff_skirt_b"],
    "17": ["sl_left_cuff_skirt_f"],
    "18": ["sl_left_cuff_skirt_b"],
    "19": ["sl_left_cuff_f"],
    "20": ["sl_left_cuff_b"],
    "21": ["left_hood"],
    "22": ["right_hood"],
    "23": ["wb_back"],
    "24": ["wb_front"],
    "25": ["pant_f_l","skirt_front",
        "skirt_panel_0"],
    "26": ["pant_b_l","skirt_back",
        "skirt_panel_1"],
    "27": ["pant_f_r","skirt_front_0",
        "ins_skirt_front_0","skirt_panel_2"],
    "28": ["pant_b_r","skirt_back_0",
        "ins_skirt_back_0","skirt_panel_3"],
    "29": ["pant_r_cuff_skirt_f","skirt_front_1",
        "ins_skirt_front_1","skirt_panel_4"],
    "29": ["pant_r_cuff_skirt_b","skirt_back_1",
        "ins_skirt_back_1","skirt_panel_5"],
    "30": ["pant_l_cuff_skirt_f","ins_skirt_front_2",
        "skirt_panel_6","skirt_front_2"],
    "31": ["pant_l_cuff_skirt_b","ins_skirt_back_2",
        "skirt_panel_7","skirt_back_2"
    ],
    "32": ["pant_l_cuff_f","ins_skirt_front_3",
        "skirt_panel_8","skirt_front_3"],
    "33": ["pant_l_cuff_b","ins_skirt_back_3",
        "skirt_panel_9","skirt_back_3"],
    "34": ["pant_r_cuff_f","ins_skirt_front_4",
        "skirt_panel_10","skirt_front_4"],
    "35": ["pant_r_cuff_b","ins_skirt_back_4",
        "skirt_panel_11","skirt_back_4"],
    "36": ["skirt_panel_12","ins_skirt_front_5"],
    "37": ["skirt_panel_13","ins_skirt_back_5"],
    "38": ["skirt_panel_14"]
}
\end{lstlisting}

\section{Comparison with Parametric Method}
\label{sec:parametric}

\begin{figure}[t]
  \centering
  \includegraphics[width=\linewidth]{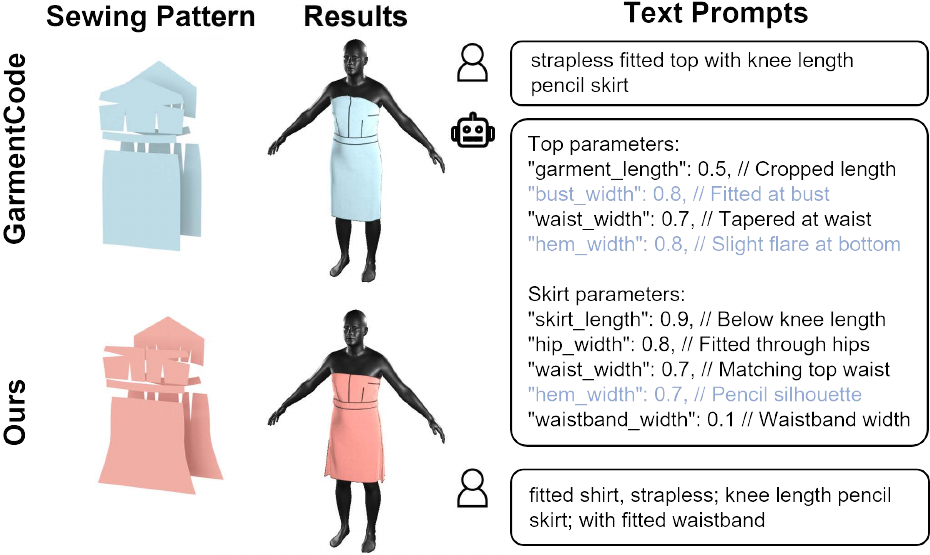}
   \caption{\textbf{Comparison with parametric method.} We present the garments and draping results for our SewingLDM and parametric method GarmentCode~\cite{korosteleva2023garmentcode}. GarmentCode needs a delicate selection of values. In contrast, our method can generate garments under intuitive conditions like natural language or sketches, which provide an easier way for garment generation.}
   \label{fig:compare_with_temp}
\end{figure}

Except for generation methods, GarmentCode~\cite{korosteleva2023garmentcode} allows users to model complex garments by selecting different parameters and producing desired sewing patterns.
However, selecting various parameters is not intuitive and requires professional knowledge of garment design, limiting its widespread promotion.
To enable the production of the desired sewing patterns through users' prompts, an easy way is to leverage the powerful ability of a large language model, like GPT-4~\cite{achiam2023gpt}.
We simply ask GPT4 to generate various values between 0 to 1 to satisfy the sewing pattern designs of~\cite{korosteleva2023garmentcode}, as illustrated in~\cref{fig:compare_with_temp}.
With the designed prompt, GPT-4 can truly provide instructions for garment design.
However, most of the generated values are not concerned with garment shape in GarmentCode~\cite{korosteleva2023garmentcode} and still require professional knowledge of garment designs and pre-defined templates.
In summary, GarmentCode~\cite{korosteleva2023garmentcode} needs indispensable manual processing to produce the desired garments.
In contrast, our SewingLDM can generate the desired garment through more intuitive conditions, \ie, text prompts and sketches, providing easier tools for garment designs and boosting daily garment production.

\begin{figure*}[t]
  \centering
  \includegraphics[width=\linewidth]{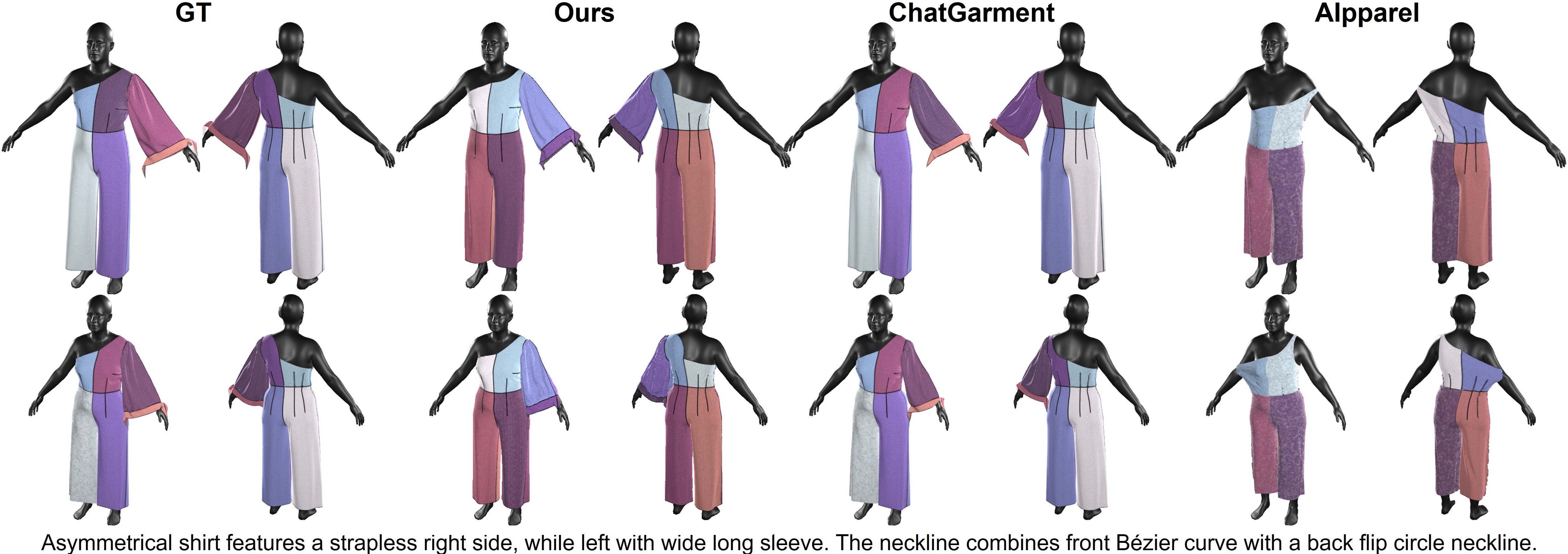}
   \caption{\textbf{Comparison of sewing pattern generation for concurrent works.} Since all the methods were trained on the same dataset, we randomly selected samples from the original dataset to evaluate and compare their performance.
   }
    \label{fig:concurrent_comparison}
\end{figure*}

\section{Concurrent Work Comparison}
\label{sec:concurrent}
For concurrent sewing pattern generation methods, both Design2GarmentCode~\cite{zhou2024design2garmentcode} and ChatGarment~\cite{bian2024chatgarment} use autoregressive models to convert inputs into rule-based parameters specially designed in GarmentCode, limiting their generalization beyond pre-defined garment categories.
AIpparel~\cite{nakayama2024aipparel}, enabling the generation of any extended sewing patterns without requiring category-specific parameterization, while it only considers the average body shape draping.
We also compared with these open-sourced methods, ChatGarment~\cite{bian2024chatgarment} and AIpparel~\cite{nakayama2024aipparel}.
Since all methods are trained on the GarmentCodeData~\cite{korosteleva2024garmentcodedata}, we randomly chose some examples in the GarmentCodeData for fair comparison, shown in~\cref{fig:concurrent_comparison}.
ChatGarment directly generates the design files and uses GarmentCode to generate appropriate sewing patterns and drape to other body shapes.
AIpparel generates sewing patterns and warps to the average body shapes and drapes to other body shapes by complete pattern scaling manually.
Compared with concurrent works, our SewingLDM shows comparable results with ChatGarment and shows much superior results with AIpparel.
Moreover, due to the body shape conditions, our method can fit various body shapes, which provides a more user-friendly approach to getting the desired made-to-measure garments.

\section{Dataset Annotation Details}
\label{sec:annotation}
Directly employing multimodal models for generating textual descriptions of images may lead to inaccuracies, as these large-scale multimodal models are incapable of fully capturing and characterizing garment-specific features, often resulting in hallucination phenomena.
To obtain precise descriptions of garment characteristics, we have established a corresponding descriptive framework based on the parameters of a parametric model, generating distinct descriptions for various garment features, which are then concatenated together to form the text annotations of garments.
The corresponding descriptions are presented as follows:

\begin{lstlisting}[language=json,firstnumber=1,basicstyle=\ttfamily\footnotesize]
{
    "sleeve length": {
        [0, 0.4]: "short",
        [0.4, 0.6]: "elbow length",
        [0.6, 0.8]: "three quarter length",
        [0.8, 1.15]: "long"
    },
    "shirt length": {
        [0, 1]: "short shirt",
        [1, 1.5]: "shirt",
        [1.5, 2.4]: "long shirt",
        [2.4, 2.8]: "knee length shirt dress",
        [2.8, 3.5]: "long shirt dress"
    },
    "neckline width": {
        [-0.5, -0.1]: "narrow",
        [-0.1, 0.5]: "normal",
        [0.5, 1]: "wide"
    },
    "neckline depth": {
        [0.3, 0.4]: "shallow",
        [0.4, 0.9]: "normal",
        [0.9, 2]: "deep"
    },
    "waist band width": {
        [0, 0.2]: "narrow waistband"
        [0.2, 0.5]: "waistband",
        [0.5, 1]: "wide waistband"
    },
    "skirt length": {
        [-0.2, 0]: "mini",
        [0, 0.25]: "short",
        [0.25, 0.4]: "above knee length",
        [0.4, 0.5]: "knee length",
        [0.5, 0.55]: "below knee length",
        [0.55, 0.65]: "midi length"
        [0.65, 0.85]: "hight ankle length",
        [0.85, 0.95]: "ankle length"
    },
    "godet width": {
        [10, 15]: "shallow",
        [15, 25]: "normal",
        [25, 50]: "deep"
    },
    "godet width": {
        [10, 15]: "tight",
        [15, 25]: "normal",
        [25, 50]: "oversized"
    },
    "pant length": {
        [0.2, 0.3]: "short",
        [0.3, 0.4]: "above knee length",
        [0.4, 0.5]: "knee length",
        [0.5, 0.55]: "below knee length",
        [0.55, 0.65]: "mid calf length"
        [0.65, 0.8]: "hight ankle length",
        [0.8, 0.9]: "full length"
    }
}
\end{lstlisting}
In addition to characterizing apparel attributes based on numerical data, we also provide corresponding descriptions according to the categories of different garments, for instance, shirt, fitted shirt, straight waistband, fitted waistband, pencil skirt, circle skirt, mermaid skirt, tail dress, etc.
Here is an example of garment description: \textit{Upper garment: asymmetrical fitted shirt, right: curve armhole, ruffled shoulder, wide short ruffled sleeve, short shrink cuffs, front wide shallow V neckline and back wide shallow V neckline; left: strapless; Lower garment: ankle length circle skirt with front right deep split}.
After getting the detailed description of the garments, we use GPT-4 to refine them, allowing for more diverse descriptions and more general prompt queries.
The prompt used to refine them is shown in~\cref{tab:prompt}.
For each garment, we utilize PiDiNet~\cite{pdc-PAMI2023}, a pre-trained edge detection network, to extract garment sketches using its open-source code and checkpoint, thereby enriching the design details of the garment.

\begin{table}[ht]\centering
\begin{minipage}{\columnwidth}\vspace{0mm}    \centering
\begin{tcolorbox} 
    \centering
    \small
     \hspace{-6mm}
\begin{itemize}[leftmargin=2mm]
\setlength{\itemsep}{1pt}
\footnotesize
\item ``\textcolor{ForestGreen}{\texttt{<prompt>}}'' Please refine the following clothing description to make it more fluent and accurate in English while strictly preserving all technical details, measurements, fabric properties, and design intent. Ensure the language is professional yet natural, suitable for a global audience. Focus on:

\begin{itemize}
    \item Clarity: Eliminate ambiguity (if applicable).
    \item Consistency: Use standardized terminology (e.g., 'midi dress' instead of 'knee-length dress').
    \item Precision: Highlight functional features (e.g., moisture-wicking, reinforced seams) without exaggeration.
    \item Flow: Restructure sentences to enhance readability while retaining all original information.
    \item SEO-friendly: Incorporate relevant keywords naturally (if applicable).
\end{itemize}

Original Text:

[Upper garment: asymmetrical fitted shirt, right: curved armhole, ruffled shoulder, wide short ruffled sleeve, short shrink cuffs, front wide shallow V neckline and back wide shallow V neckline; left: strapless; Lower garment: ankle length circle skirt with front right deep split.]

Provide the revised version.
\item This asymmetric tailored shirt features a dual design: Right side: Contoured armhole with ruffled detailing at the shoulder, voluminous short ruffled sleeves, elasticized short cuffs, and dual wide shallow V-necklines (front and back). Left side: A strapless silhouette for bold contrast. Paired with an ankle-length full-circle skirt, the ensemble includes a vented high slit at the right front panel for dynamic movement and versatility.
\end{itemize}
\end{tcolorbox}
\vspace{-1em}
\caption{\textbf{Prompt to refine the dataset annotations.}}
    \label{tab:prompt}
\end{minipage}
\end{table}

\section{Extended Vector v.s Origin Vector}
\label{sec:extend_rep}
\begin{figure}[t]
  \centering
  \includegraphics[width=\linewidth]{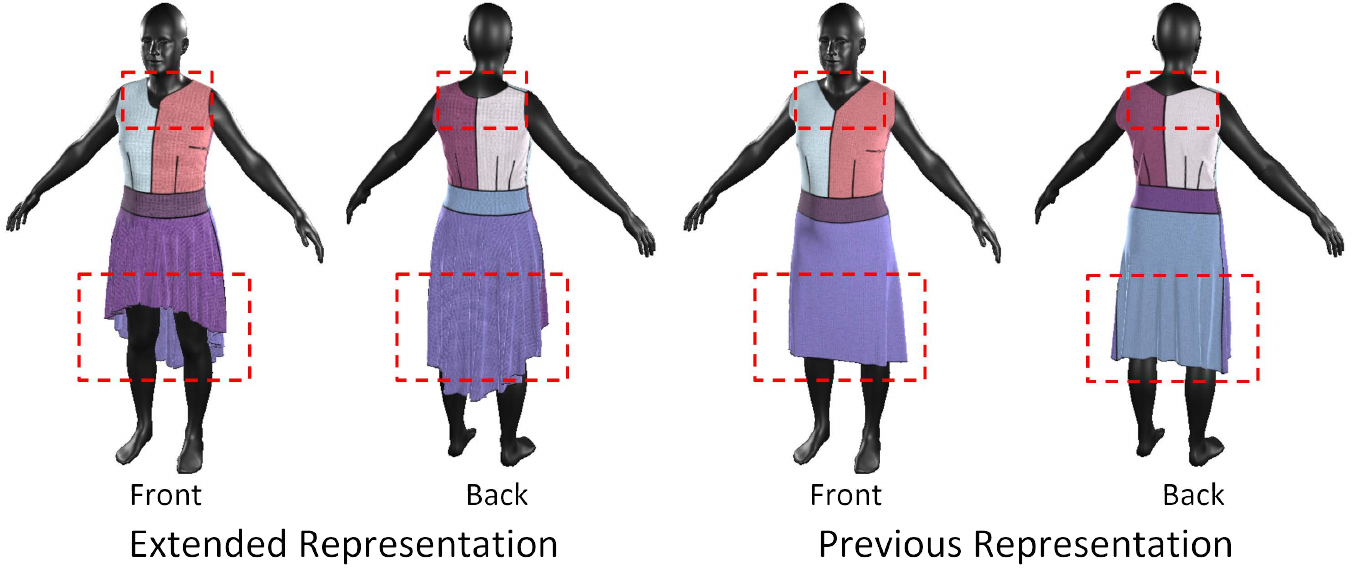}
   \caption{\textbf{Visulization of different representation.} We compare the extended representation with the original representation.}
   \vspace{-1em}
   \label{fig:representation_ablation}
\end{figure}
We also conduct the visualization between the extended representation and original representation to reveal significant advantages in design versatility of the extended representation shown in~\cref{fig:representation_ablation}.
With extended representation, our model can generate more diverse garments and more fashion designs, such as the curve neckline and the tail dress.
In contrast, the original representation restricts the model to producing only simple garments, failing to capture complex design features and resulting in a notable loss of detail.
To address these limitations, the integration of the GarmentCodeData is essential, as it injects detailed knowledge of complex sewing patterns into the generation model.
This advancement overcomes the constraints of previous methods, which were unable to represent the extended vector effectively due to more than 30k tokens consumption under current GPU capabilities.

\section{One-step Training v.s Two-step Training}
\label{sec:ablation}
\begin{figure}[t]
  \centering
  \includegraphics[width=\linewidth]{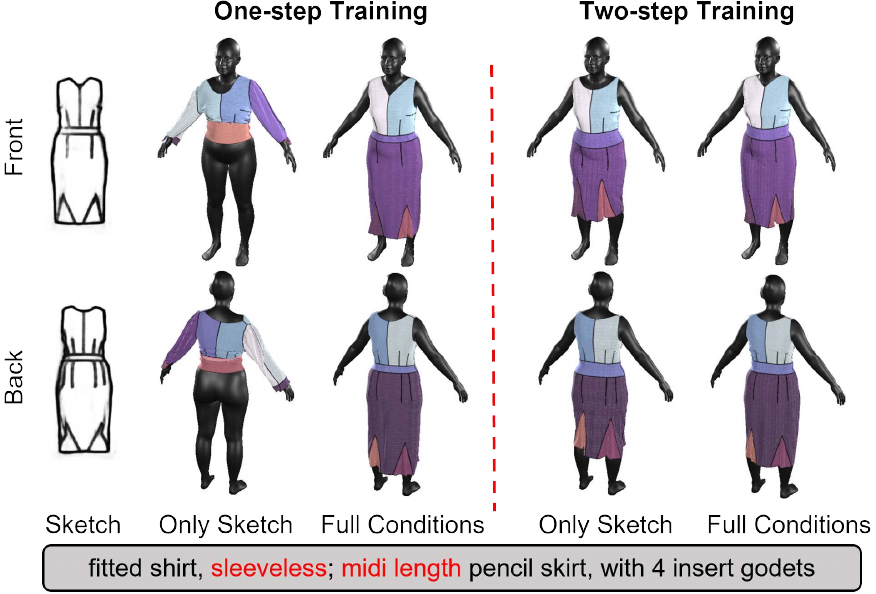}
  \vspace{-1.5em}
   \caption{\textbf{Ablation on the training strategy.} One-step training shows an unbalance between the multi-modal conditions, failing under only the sketch. In contrast, two-step training helps to faithfully generate the ideal garments under only sketch conditions.}
   \vspace{-1.5em}
   \label{fig:ablation}
\end{figure}
We additionally conduct an ablation study on the training strategy.
One-step training is unable to balance the multi-modal conditions, so that fails to generate the corresponding garment through only the sketch condition.
As shown in~\cref{fig:ablation}, the generated garment loses its midi-length pencil skirt, failing to generate the desired garment.
In contrast, the model under two-step training can faithfully generate the corresponding garment with the sketch only, which contains both the sleeveless fitted shirt and the corresponding midi-length pencil skirt.
Therefore, two-step training can more effectively inject the sketch conditions into the diffusion model and provide additional control of garment designs, enabling wider usage of our SewingLDM.
Moreover, combined with full conditions of text and sketch, SewingLDM can provide more precise control over desired garment generations, meeting users' requirements.

\section{User Study Details}
\label{sec:user_study}
\begin{figure}[t]
  \centering
  \includegraphics[width=\linewidth]{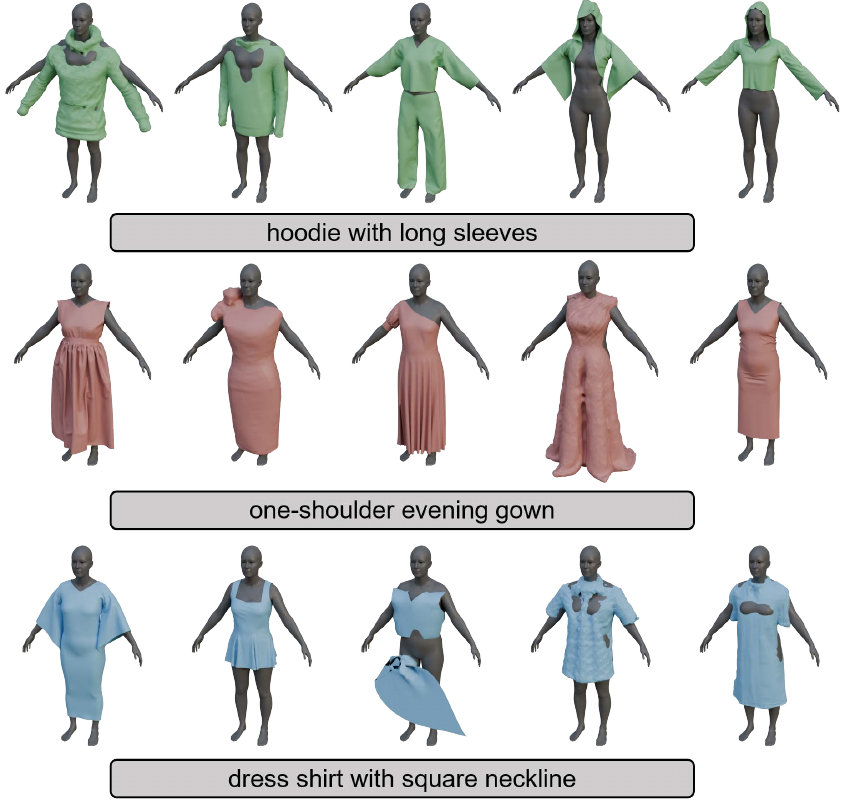}
  \vspace{-1em}
   \caption{\textbf{User study examples.} We present 3 user study examples with random shuffled results.}
   \vspace{-1em}
   \label{fig:user_study}
\end{figure}
To ensure a fair and objective evaluation of our method compared to other methods, we randomly shuffle the results generated by different methods.
Each result is paired with a corresponding textual description, and volunteers are asked to rate the results with a score of $1-5$ based on the consistency between the results and the texts, as well as the fitness between the clothes and the human bodies.
Additionally, we provide 3 supplementary examples as shown in~\cref{fig:user_study}.

\section{Use Case}
\label{sec:use_case}
\begin{figure}[t]
  \centering
  \includegraphics[width=\linewidth]{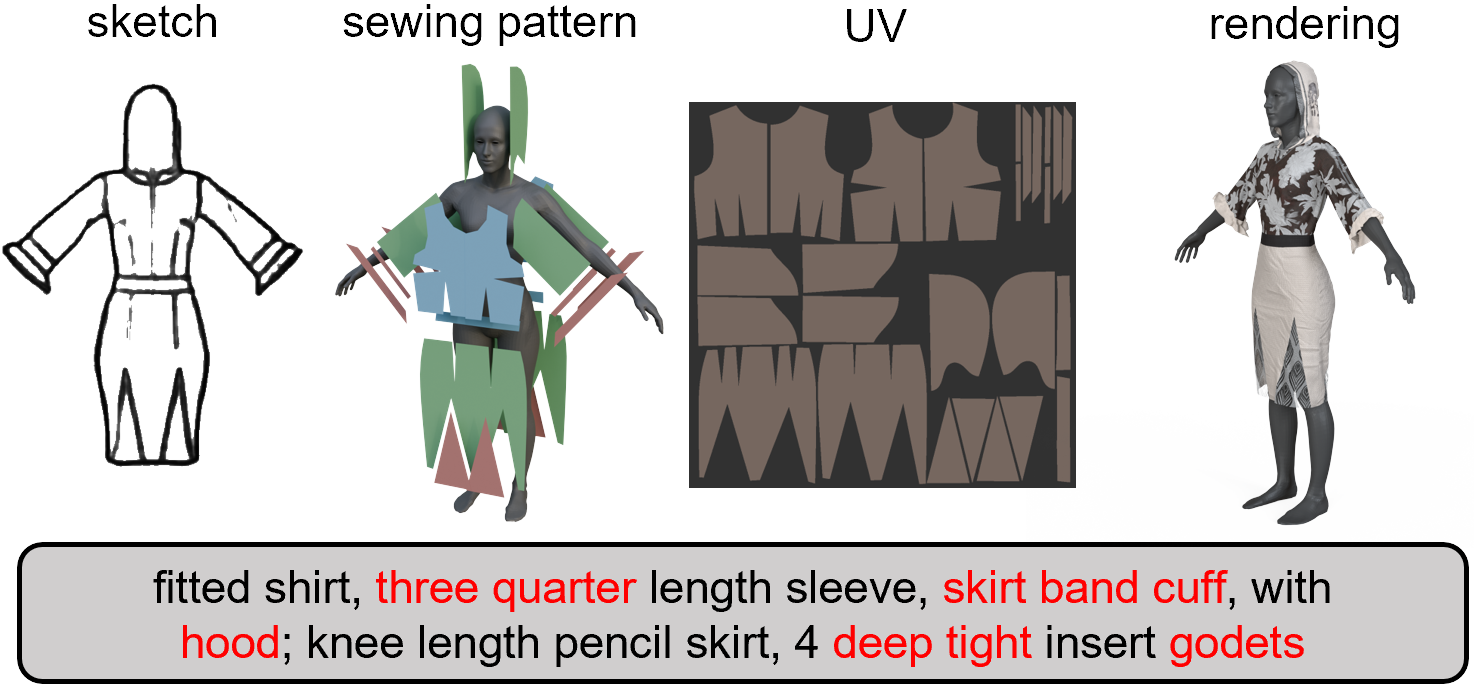}
   \caption{\textbf{Use case.} We present an example of an extremely complex sewing pattern generation. With the generated sewing pattern, we can easily paint the UV to produce visually appealing results.}
   \vspace{-1.5em}
   \label{fig:user_case}
\end{figure}

Our SewingLDM is capable of generating intricately detailed clothing, meeting the current artistic demands for garment design across a wide range of styles, significantly advancing fashion garment design, and supporting everyday users in obtaining apparel tailored precisely to their needs.
To demonstrate our superiority in garment generation, we present an extremely complex example of a sewing pattern in~\cref{fig:user_case}.
With the detailed textual description and garment sketch, our method faithfully generates the complex sewing pattern, \eg, skirt band cuff, hood, and godets, which significantly helps the artist in creating fantastic texture in UV space, \eg, laces, leather pants, and hat brim.

\section{Qualitative Results}
\label{sec:qualitave}
\begin{figure*}[t]
  \centering
  \includegraphics[width=\linewidth]{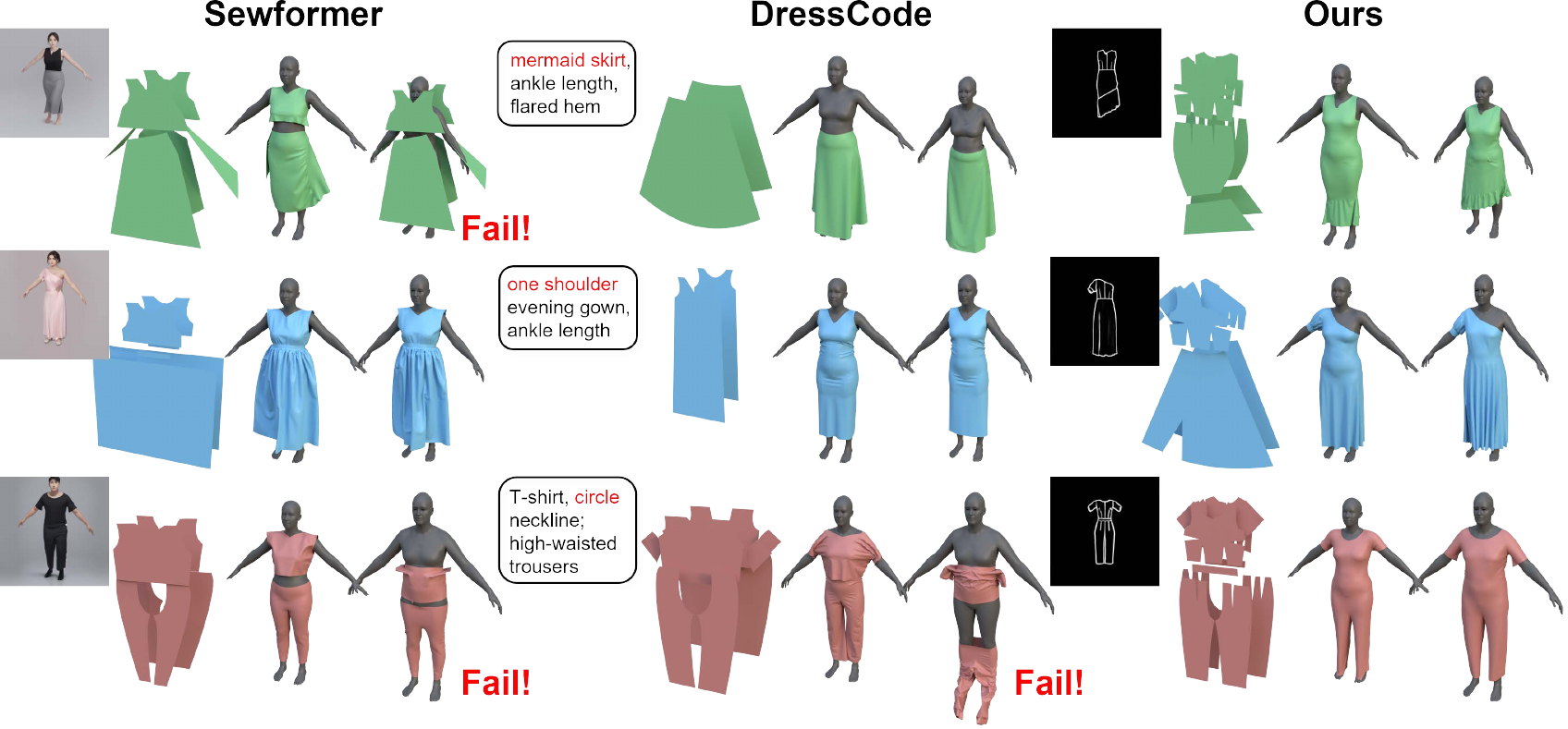}
  \vspace{-1.5em}
   \caption{\textbf{Additional results for in-domain sketches.} We present the in-domain comparison with baseline methods.}
   \vspace{-0.5em}
   \label{fig:in_domain}
\end{figure*}

\begin{figure*}[t]
  \centering
  \includegraphics[width=\linewidth]{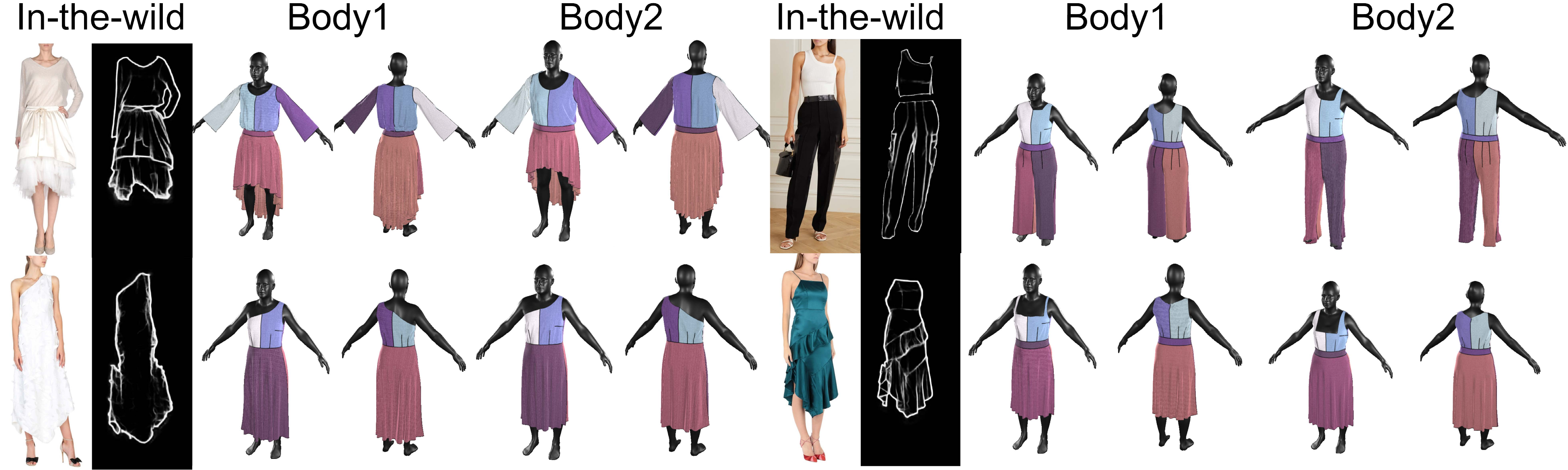}
   \vspace{-1.5em}
   \caption{\textbf{Additional results for in-the-wild sketches.} We also provide more results for in-the-wild sketches.}
   \vspace{-1.5em}
   \label{fig:sub_in_the_wild}
\end{figure*}

\begin{figure}[t]
  \centering
  \includegraphics[width=\linewidth]{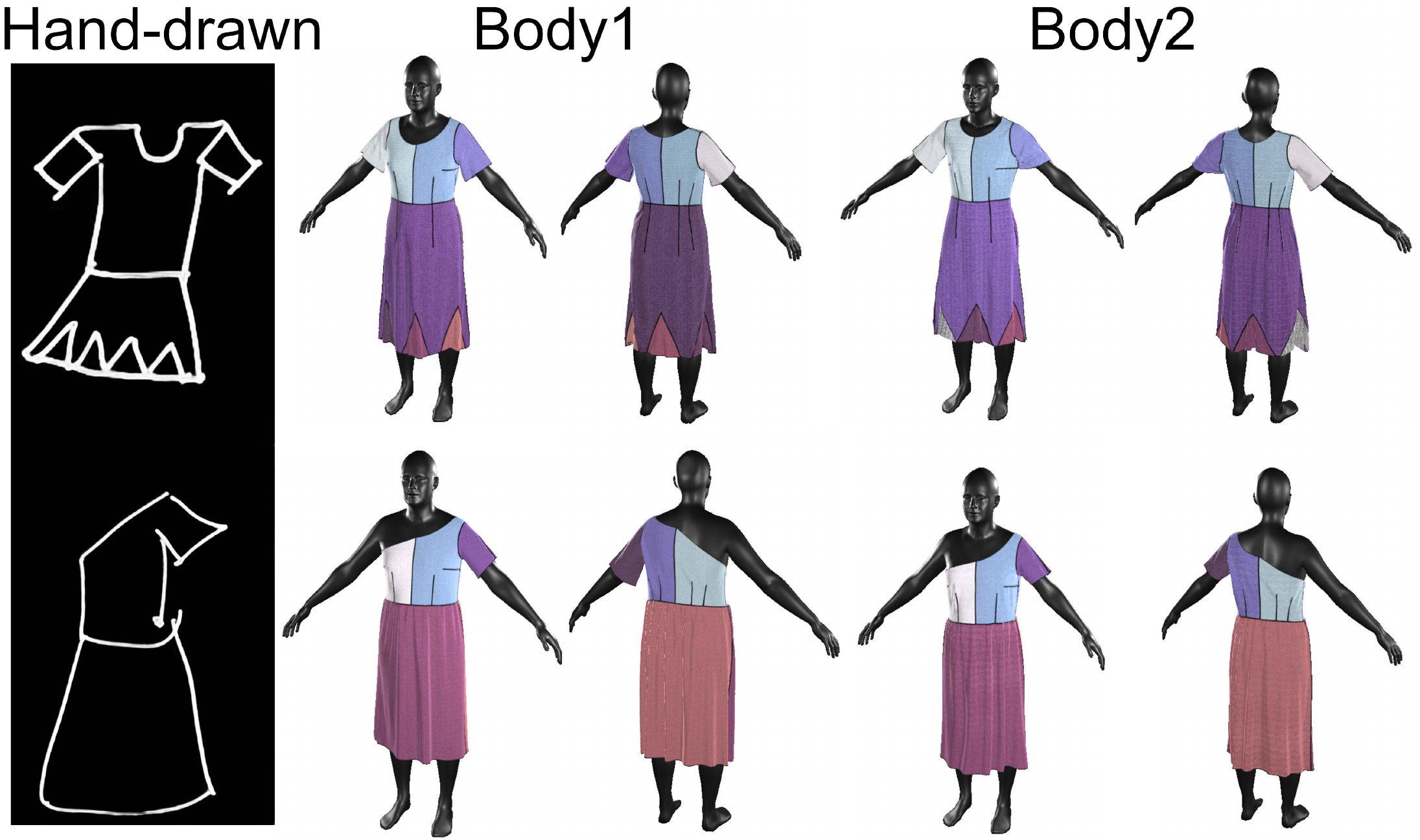}
   \caption{\textbf{Additional results for hand-drawn sketches.}}
   \vspace{-1.5em}
   \label{fig:hand_drawn}
\end{figure}

For in-domain garment sketches, our method achieves a more granular representation that closely adheres to the sketch contours.
We have also provided a comparative analysis between our approach and other baseline methods within the same domain, as illustrated in~\cref{fig:in_domain}. Furthermore, to demonstrate the efficacy of our method, we have conducted additional validation using out-of-domain data that are collected from the Multimodal Garment Designer or hand-drawn, as depicted in~\cref{fig:sub_in_the_wild} and~\cref{fig:hand_drawn}.

\begin{figure*}[t]
  \centering
  \includegraphics[width=0.9\linewidth]{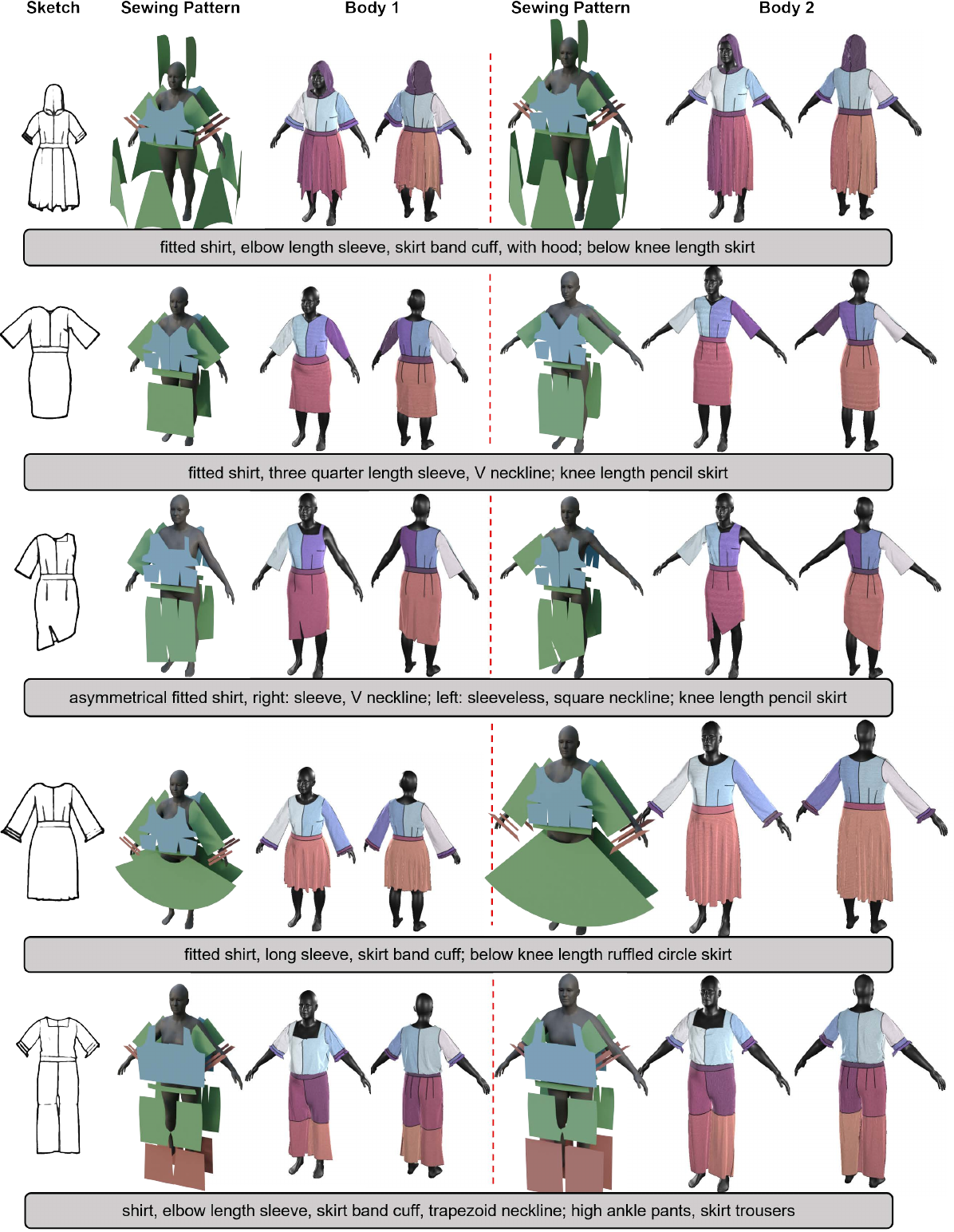}
   \caption{\textbf{Additional results for various body shapes.} We present identical garment designs tailored for two distinct body types, encompassing a spectrum of heights and body compositions, to demonstrate the effectiveness of our SewingLDM across diverse bodies.}
   \label{fig:sub_body}
\end{figure*}

\begin{figure*}[t]
  \centering
  \includegraphics[width=0.9\linewidth]{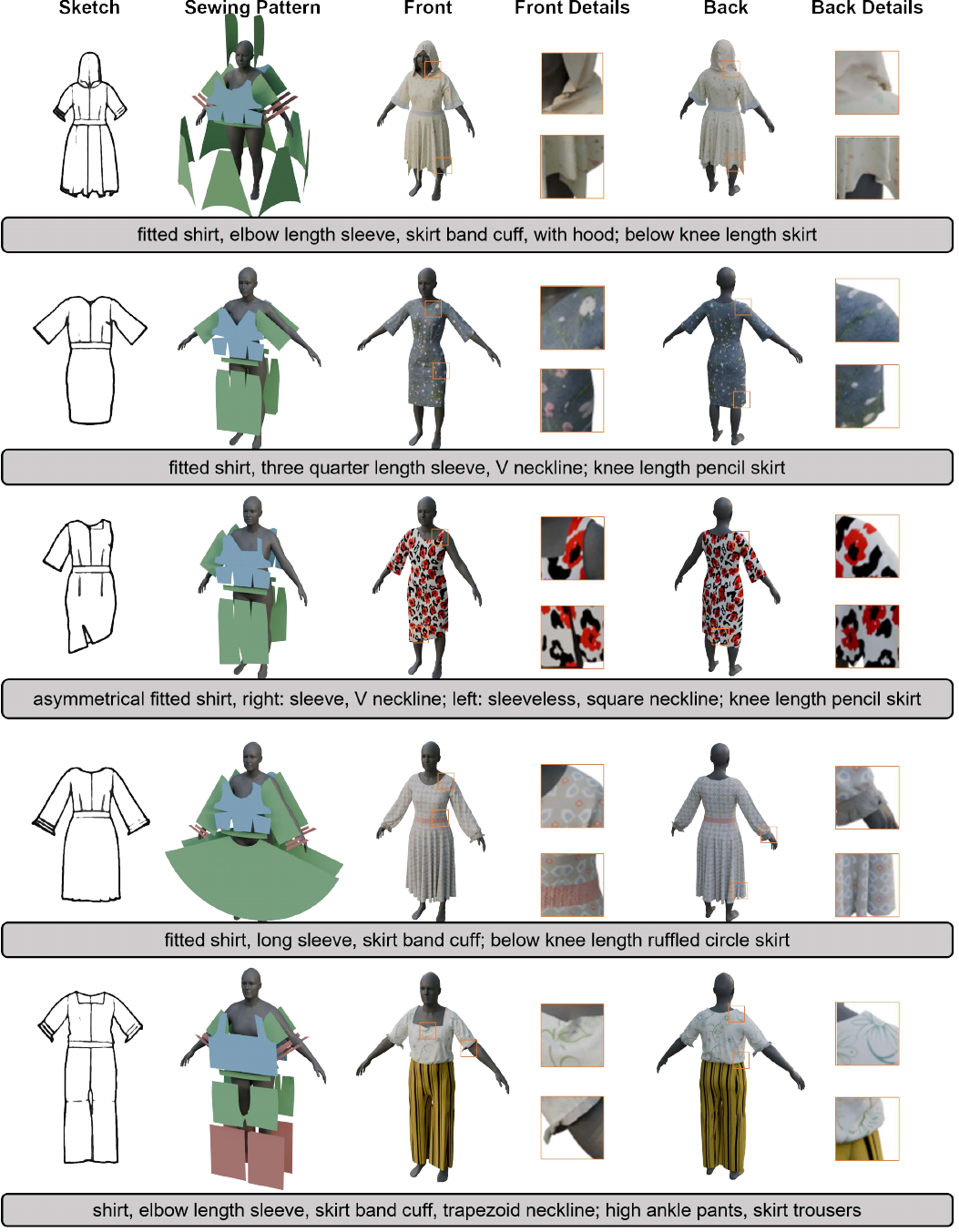}
   \caption{\textbf{Additional qualitative results.} By integrating Physically Based Rendering (PBR) textures, our generated outputs achieve visually compelling rendering effects, particularly for a wide range of intricate garment designs.}
   \label{fig:sub_image}
\end{figure*}

We additionally provide garments tailored to a wide range of body shapes, spanning variations such as short to tall and slim to broad. As illustrated in~\cref{fig:sub_body}, our approach enables the creation of garments specifically adapted to different body types.
Furthermore, the simulated garments are enriched with physically based rendering (PBR) textures, either generated by DressCode or designed using the Substance 3D Painter software~\cite{sp}, culminating in visually compelling garment representations as shown in~\cref{fig:sub_image}.

\end{document}